\documentclass[pdflatex,sn-mathphys-num,iicol]{sn-jnl}

\usepackage{graphicx}%
\usepackage{multirow}%
\usepackage{amsmath,amssymb,amsfonts}%
\usepackage{amsthm}%
\usepackage{mathrsfs}%
\usepackage[title]{appendix}%
\usepackage[table]{xcolor}%
\usepackage{ragged2e}%
\usepackage{textcomp}%
\usepackage{manyfoot}%
\usepackage{booktabs}%
\usepackage{algorithm}%
\usepackage{algorithmicx}%
\usepackage{algpseudocode}%
\usepackage{listings}%
\usepackage{placeins}%

\theoremstyle{thmstyleone}%
\theoremstyle{thmstyletwo}%

\theoremstyle{thmstylethree}%

\begin{document}

\title[Thermal Topology Collapse]{Thermal Topology Collapse: Universal Physical Patch Attacks on Infrared Vision Systems}


\author[1]{\fnm{Chengyin} \sur{Hu}}\email{cyhu@cupk.edu.cn}

\author[1]{\fnm{Yikun} \sur{Guo}}\email{gyk666@st.cupk.edu.cn}

\author[1]{\fnm{Yuxian} \sur{Dong}}\email{854531750@st.cupk.edu.cn}

\author[1]{\fnm{Qike} \sur{Zhang}}\email{zhangqike@st.cupk.edu.cn}

\author[2]{\fnm{Kalibinuer} \sur{Tiliwalidi}}\email{kalibinur@gdut.edu.cn}

\author[1]{\fnm{Yiwei} \sur{Wei}}\email{weiyiwei@cupk.edu.cn}

\author[1]{\fnm{Haitao} \sur{Shi}}\email{shinian@cupk.edu.cn}

\author[3]{\fnm{Jiujiang} \sur{Guo}}\email{gjiujiang@163.com}

\author*[4]{\fnm{Jiahuan} \sur{Long}}\email{jiahuanlong@sjtu.edu.cn}

\affil[1]{\orgname{China University of Petroleum-Beijing at Karamay}, \orgaddress{\street{No. 355, Anding Road}, \city{Karamay}, \postcode{834000}, \state{Xinjiang}, \country{China}}}

\affil[2]{\orgname{Guangdong University of Technology}, \orgaddress{\city{Guangzhou}, \state{Guangdong}, \country{China}}}

\affil[3]{\orgname{Tianjin University}, \orgaddress{\city{Tianjin}, \country{China}}}

\affil*[4]{\orgname{Shenzhen University}, \orgaddress{\city{Shenzhen}, \country{China}}}


\abstract{Infrared pedestrian detectors are increasingly deployed in all-weather perception systems, but their robustness against physical adversarial attacks remains insufficiently understood. Existing infrared physical attacks are mostly instance-specific, requiring perturbations to be optimized for particular samples, poses, or scenes, which limits scalability under changing deployment conditions. This paper proposes Universal Physical Patch Attack (UPPA), a universal cold-patch framework for infrared pedestrian detection. UPPA is built on the observation that infrared physical perturbations should exploit smooth, low-frequency thermal structures rather than visible-light texture patterns. It represents the attack carrier as topology-constrained B\'{e}zier Curved-Blocks, providing a compact and manufacturable geometric parameterization, and optimizes one shared perturbation with Particle Swarm Optimization (PSO) under Thin Plate Spline (TPS) deformation and Expectation over Transformation (EOT) imaging transformations. The optimized pattern is then deployed as wearable cold patches without sample-specific re-optimization during deployment. Experiments on five infrared datasets and nine pedestrian detectors show consistent digital attack performance, strong cross-dataset and cross-model transferability, and a 92.59\% attack success rate in real-world physical experiments. Ablation, visualization, and defense analyses show that Curved-Blocks disrupt pedestrian thermal feature aggregation and remain difficult for image-restoration-style defenses to remove. These results reveal a practical universal physical vulnerability in current infrared pedestrian detection systems and provide a benchmark for robustness evaluation.}




\maketitle

\section{Introduction}\label{sec1}

Deep Neural Networks (DNNs) have become the dominant technical foundation for modern visual perception, achieving remarkable progress in image classification~\cite{he2016deep} and object detection~\cite{redmon2018yolov3}. In safety-critical applications such as autonomous driving, perimeter surveillance, and night-time pedestrian monitoring, reliable object detection must remain effective under poor illumination, adverse weather, and large appearance variations. Visible-light sensors often suffer from severe signal degradation in such uncontrolled environments~\cite{sakaridis2021acdc}, whereas infrared thermal imaging captures the radiation distribution of objects and therefore provides an important complementary modality for all-weather perception~\cite{hwang2015multispectral}. As infrared pedestrian detectors are increasingly integrated into practical vision systems, understanding their robustness is no longer only a theoretical concern but also a prerequisite for trustworthy deployment. Recent evidence suggests that this robustness issue appears in both digital and physical settings. For example, attention-guided sparse attacks reveal the digital vulnerability of DNN-based visual models~\cite{li2023adaptive}, while infrared adversarial stickers show that thermal perception systems can also be manipulated by realizable physical patterns~\cite{zhu2024infraredstickers}.

\begin{figure*}[!t]
    \centering
    \includegraphics[width=\textwidth]{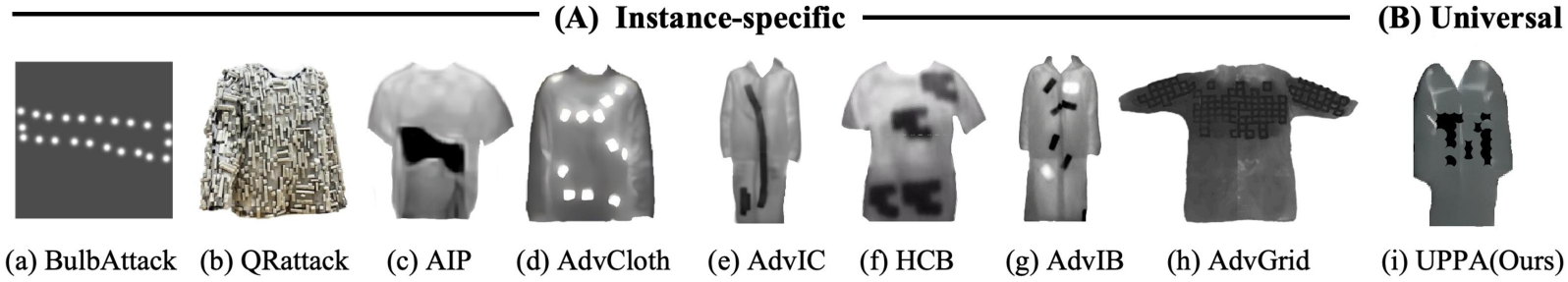}
    \caption{Infrared physical attack methods against pedestrian detectors. (A) Prior instance-specific methods, including BulbAttack~\cite{zhu2021fooling}, QRattack~\cite{zhu2022infrared}, AIP~\cite{wei2023physically}, AdvCloth~\cite{zhu2023hiding}, HCB~\cite{wei2023hotcold}, AdvIB~\cite{hu2024adversarial-blocks}, AdvIC~\cite{hu2024adversarial-curves}, and AdvGrid~\cite{tiliwalidi2025advgrid}; (B) Universal (UPPA, ours).}
    \label{fig:infrared_attacks_evolution}
\end{figure*}

Research on physical adversarial attacks has consequently expanded from the visible spectrum to the infrared domain, where the attack carrier must obey the imaging physics of thermal radiation. Existing infrared physical attacks, as illustrated in Fig.~\ref{fig:infrared_attacks_evolution}, first demonstrated feasibility with active heat-source designs. BulbAttack~\cite{zhu2021fooling} uses small bulbs to generate salient thermal responses, and AdvCloth~\cite{zhu2023hiding} embeds heating films into clothing to produce adversarial thermal patterns. Later work moved toward passive and more deployable media. HCB~\cite{wei2023hotcold} introduces wearable hot--cold blocks for black-box thermal attacks, AdvIB~\cite{hu2024adversarial-blocks} studies multi-view adversarial infrared blocks, AdvIC~\cite{hu2024adversarial-curves} replaces rigid blocks with curve-like structures, and AdvGrid~\cite{tiliwalidi2025advgrid} explores grid-style physical optimization. This development has substantially improved the feasibility of attacking infrared pedestrian detectors in the physical world. Meanwhile, universal adversarial perturbations have been extensively investigated in the visible domain, where hard-label black-box universal patches~\cite{tao2023hard} show that a single perturbation can sometimes exploit shared model vulnerabilities across many input instances. These two lines of work suggest a natural but still underexplored question: can one construct a universal physical perturbation for infrared pedestrian detection that is both adversarially effective and physically realizable?

\begin{table*}[t]
  \centering
  \caption{\protect\justifying Comparison of representative infrared physical attack methods against pedestrian detectors. Stealthiness: human-perceptual concealment without auxiliary infrared sensors; Paradigm: whether the perturbation is optimized for each individual instance or learned once as a universal pattern; Real-time: direct physical deployment of the digitally optimized perturbation without per-input re-optimization.}
  \label{tab:intro_method_comparison}
  \footnotesize
  \renewcommand{\arraystretch}{1.15}
  \setlength{\tabcolsep}{7pt}
  \begin{tabular*}{\textwidth}{@{\extracolsep{\fill}} l l c c c c @{}}
    \toprule
    Method & Perturbation & Stealthiness & Scenario & Paradigm & Real-time \\
    \midrule
    BulbAttack~\cite{zhu2021fooling} & Light bulbs & \textcolor{red}{$\times$} & White-box & Instance-specific & \textcolor{red}{$\times$} \\
    QRattack~\cite{zhu2022infrared} & Aerogels & \textcolor{red}{$\times$} & White-box & Instance-specific & \textcolor{red}{$\times$} \\
    AIP~\cite{wei2023physically} & Aerogels & \textcolor{red}{$\times$} & White-box & Instance-specific & \textcolor{red}{$\times$} \\
    AdvCloth~\cite{zhu2023hiding} & Electric heating films & \textcolor{green!60!black}{$\checkmark$} & White-box & Instance-specific & \textcolor{red}{$\times$} \\
    HCB~\cite{wei2023hotcold} & Hot and cold pastes & \textcolor{green!60!black}{$\checkmark$} & Black-box & Instance-specific & \textcolor{red}{$\times$} \\
    AdvIB~\cite{hu2024adversarial-blocks} & Hot and cold pastes & \textcolor{green!60!black}{$\checkmark$} & Black-box & Instance-specific & \textcolor{red}{$\times$} \\
    AdvIC~\cite{hu2024adversarial-curves} & Cold pastes & \textcolor{green!60!black}{$\checkmark$} & Black-box & Instance-specific & \textcolor{red}{$\times$} \\
    AdvGrid~\cite{tiliwalidi2025advgrid} & Cold pastes & \textcolor{green!60!black}{$\checkmark$} & Black-box & Instance-specific & \textcolor{red}{$\times$} \\
    UPPA (Ours) & Cold pastes & \textcolor{green!60!black}{$\checkmark$} & Black-box & Universal & \textcolor{green!60!black}{$\checkmark$} \\
    \bottomrule
  \end{tabular*}
\end{table*}

Answering this question is difficult because infrared physical attacks face constraints that are fundamentally different from those in visible-light attacks. Visible-domain universal patches usually benefit from rich color channels, high-frequency textures, and pixel-level pattern freedom, but these assumptions do not hold for infrared imaging. Thermal diffusion, low spatial resolution, single-channel thermal intensity, and the material properties of cold or hot media make it hard to reproduce discrete high-frequency perturbations in the physical world. Under these constraints, existing infrared physical attacks often maintain attack efficacy by adapting the perturbation to a particular pedestrian pose, scale, or frame, leading to a ``single-sample-single-optimization'' mechanism. For example, HCB~\cite{wei2023hotcold} relies on hot--cold block placement, AdvIB~\cite{hu2024adversarial-blocks} further optimizes adversarial infrared blocks across views, and AdvGrid~\cite{tiliwalidi2025advgrid} organizes the perturbation as a grid-like pattern. This instance-specific paradigm introduces two major bottlenecks: it requires repeated optimization or adjustment before deployment, and it tends to overfit static observations rather than remaining robust under non-rigid clothing deformation, body motion, sensor noise, and scene changes. We summarize these differences in Table~\ref{tab:intro_method_comparison} and highlight the lack of a stealthy, black-box, universal, and directly deployable infrared physical attack.

Motivated by this gap, we propose Universal Physical Patch Attack (UPPA), a universal physical patch attack framework for infrared pedestrian detectors. UPPA is designed around the observation that an infrared physical perturbation should not imitate the high-frequency texture paradigm of visible-light attacks; instead, it should encode smooth, low-frequency, and manufacturable thermal structures that can reveal common vulnerabilities of detector feature representations. Specifically, UPPA represents adversarial perturbations with topologically constrained B\'{e}zier Curved-Blocks, which compress the high-dimensional pixel search space into a compact geometric parameter space while preserving continuous boundaries suitable for cold-patch realization. To search for an instance-agnostic perturbation under black-box conditions, UPPA adopts Particle Swarm Optimization (PSO)~\cite{kennedy1995particle} over the entire dataset and incorporates both Thin Plate Spline (TPS)~\cite{bookstein1989principal} and Expectation over Transformation (EOT)~\cite{athalye2018synthesizing} to model non-rigid deformation and physical imaging variations. The optimized digital perturbation is then materialized as wearable cold patches, producing a smooth low-temperature distribution that conforms to infrared thermal characteristics and can be deployed without sample-specific re-optimization during deployment.

The main contributions of this paper are summarized as follows:

\begin{itemize}
  \item[-] \textbf{Universal infrared physical attack.} To the best of our knowledge, UPPA is the first universal physical attack method for infrared pedestrian detection. It learns a single perturbation over the entire dataset, and the digitally optimized perturbation can be directly deployed in the physical world without repeated optimization during deployment, thereby reducing computational and deployment costs.
  \item[-] \textbf{Physically realizable cold-patch design.} We model the attack carrier as topology-constrained B\'{e}zier Curved-Blocks, producing smooth low-frequency boundaries that are compatible with infrared imaging physics and practical cold-patch fabrication.
  \item[-] \textbf{Black-box robust optimization.} We formulate perturbation generation as a compact geometric search problem and optimize it with PSO under TPS and EOT-based transformation modeling, improving robustness to non-rigid deformation and physical imaging variations.
  \item[-] \textbf{Comprehensive empirical evidence.} We validate UPPA across multiple detectors, datasets, transfer settings, physical distances, and defense mechanisms, showing strong attack effectiveness, cross-domain generalization, and resistance to image-restoration-style defenses.
\end{itemize}

The remainder of this paper is organized as follows. Section~\ref{sec2} reviews physical adversarial attacks in the visible and infrared domains, as well as universal adversarial attack methods. Section~\ref{sec3} presents the UPPA framework, including the problem formulation, B\'{e}zier Curved-Block parameterization, topological constraints, and PSO-based black-box optimization. Section~\ref{sec:experiments} reports the experimental settings and evaluates UPPA in digital and physical scenarios. The subsequent discussion analyzes transferability, ablation results, attack mechanisms, and adversarial defenses, followed by the concluding remarks.

\section{Related Works}\label{sec2}

\subsection{Physical Attacks in the Visible Domain}

Physical adversarial attacks in the visible domain aim to evaluate whether deep visual models remain reliable when adversarial perturbations are manufactured and observed in the real world~\cite{eykholt2018robust}. Unlike digital attacks, physical attacks must remain effective under viewpoint changes, illumination variation, sensor noise, object motion, and background clutter. For pedestrian detection, Thys et al.~\cite{thys2019fooling} demonstrated that a localized adversarial patch can suppress the response of person detectors, establishing a practical attack setting in which the perturbation is attached to or worn by the target. This line of work has since shifted from proving physical feasibility to improving robustness, transferability, and visual naturalness.

Subsequent visible-domain methods model the physical transformation process more explicitly according to the target object. For pedestrian or person-related attacks, adversarial texture methods study wearable perturbations for fooling person detectors~\cite{hu2022adversarial}, while invisibility-cloak-style attacks examine real-world attacks on object detectors~\cite{wu2020making}. Dynamic adversarial patches further improve attack plausibility for person detectors~\cite{guesmi2024dap}. For rigid targets such as vehicles, researchers have explored several complementary directions: CNCA emphasizes customizable and natural camouflage generation~\cite{lyu2024cnca}, RAUCA improves robust and accurate physical camouflage~\cite{10.5555/3692070.3694638}, gradient-reweighted optimization strengthens detection evasion~\cite{liang2025gradient}, and 3D Gaussian Splatting supports multi-view camouflage generation~\cite{lou2025pga}. Other studies reduce human perceptibility through naturalistic patch synthesis~\cite{hu2021naturalistic} or visually legitimate adversarial designs~\cite{tan2021legitimate}. These methods provide important insights into physical robustness modeling, but they largely rely on RGB-specific degrees of freedom, including color contrast, high-frequency textures, and fine-grained spatial patterns. Since infrared imaging records thermal radiation rather than reflected visible light, such texture-driven attack mechanisms cannot be directly reproduced by cold or hot physical media. Therefore, visible-domain physical attacks offer useful transformation and deployment principles, but they do not resolve the modality-specific constraints of infrared physical attack generation.

\subsection{Physical Attacks in the Infrared Domain}

Infrared physical attacks must manipulate the thermal appearance of a target rather than its visible texture, and existing methods have therefore evolved around different physical media, access assumptions, and deployment constraints (Fig.~\ref{fig:infrared_attacks_evolution}). Early studies mainly rely on active thermal sources. Bulbs Attack~\cite{zhu2021fooling} uses light bulbs to create salient infrared responses, and AdvCloth~\cite{zhu2023hiding} employs heating films to generate adversarial thermal patterns. These methods demonstrate that infrared detectors can be fooled in the physical world, but active heat sources require power supply and careful device control, which weakens long-term deployability and visible-spectrum stealthiness.

To reduce deployment burden, later infrared attacks turn to passive cold or hot media that can be attached to clothing or placed on the target surface. QRattack~\cite{zhu2022infrared} and AIP~\cite{wei2023physically} exploit thermal insulation materials to construct pixelated perturbation patterns under white-box settings. HCB~\cite{wei2023hotcold} further introduces a hot--cold patch mechanism to improve attack effectiveness in black-box scenarios. More recent methods refine the perturbation geometry: AdvIB~\cite{hu2024adversarial-blocks} models adversarial infrared blocks, AdvIC~\cite{hu2024adversarial-curves} introduces curve-based perturbation structures, and AdvGrid~\cite{tiliwalidi2025advgrid} explores grid-style optimization for physical infrared attacks. Together, these works substantially advance infrared adversarial research from active interference toward more practical passive perturbations.

However, most infrared physical attacks remain instance-specific: their perturbations are optimized for a particular pedestrian, pose, scale, or frame, and often lose effectiveness under motion, clothing deformation, viewpoint changes, or scene shifts. This paradigm requires repeated optimization during deployment and may even demand remanufacturing or repositioning of the physical medium. In addition, rigid blocks, grids, and local curves provide limited geometric expressiveness under infrared thermodynamics and low spatial resolution. A practical infrared attack therefore needs to be universal across instances, thermally realizable, and robust to non-rigid physical transformations, which existing methods do not fully achieve.

\subsection{Universal Adversarial Attacks}

Universal adversarial attacks seek a single perturbation that generalizes across many samples, revealing shared vulnerable directions in deep models. Early data-independent methods show that universal perturbations can be generated by exploiting feature statistics rather than optimizing each image separately~\cite{BMVC2017_30}. More recent data-free approaches further introduce pseudo-semantic priors to improve universal perturbation quality without relying on labeled training data~\cite{lee2025data}. This paradigm has also been extended to physical and black-box settings. CDUPatch studies color-driven universal patches for dual-modal visible-infrared detectors~\cite{long2025cdupatch}, while hard-label black-box universal patch attacks reduce the amount of model feedback needed during optimization~\cite{tao2023hard}. For physical attacks, universality is especially valuable because the perturbation can be optimized once and reused across different instances without sample-specific re-optimization.

However, most universal attack mechanisms are designed for visible-light images and rely on visible-spectrum freedoms such as color diversity, local texture detail, and high-frequency patterns. These assumptions do not hold in infrared imaging, where thermal radiation, diffusion, and low sensor resolution favor smooth low-frequency temperature distributions. Directly transferring visible-domain universal attacks to infrared physical scenarios therefore leaves a gap between digital optimization and physical realization.

Our work addresses this missing universal infrared attack setting. Unlike instance-specific infrared attacks, UPPA learns one shared perturbation over the dataset, which can be reused during deployment without re-optimization. Unlike visible-domain universal attacks, UPPA uses topologically constrained B\'{e}zier Curved-Blocks and optimizes this compact geometric representation under TPS and EOT-based physical transformations, respecting the smoothness, manufacturability, and deformation constraints of thermal media.

\section{Methods}\label{sec3}

\begin{figure*}[t]
  \centering
  \includegraphics[width=\textwidth]{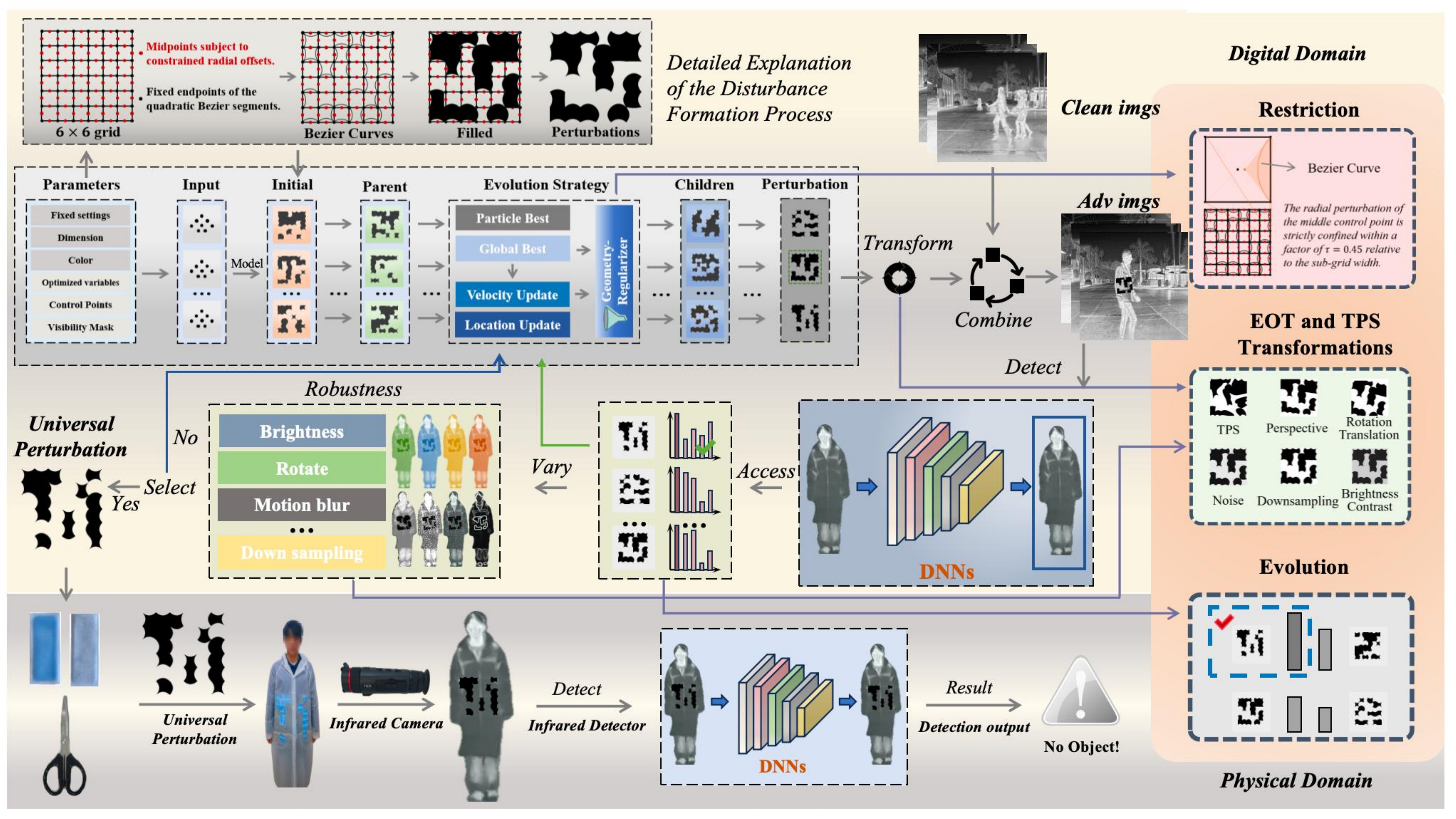}
  \caption{Overall framework of UPPA. The pipeline consists of three components: digital optimization of topology-constrained B\'{e}zier Curved-Blocks using PSO-based black-box optimization, robustness modeling with TPS and EOT transformations, and physical deployment as wearable cold patches.}
  \label{fig:framework_overview}
\end{figure*}

This section presents Universal Physical Patch Attack (UPPA), a universal physical attack framework for infrared pedestrian detection. The central idea is to learn a single thermally realizable perturbation pattern that can be shared across pedestrian instances, rather than optimizing a new pattern for each image or pose. As shown in Fig.~\ref{fig:framework_overview}, UPPA is organized into three connected components. First, the digital optimization component represents the perturbation as topology-constrained B\'{e}zier Curved-Blocks and searches their shared geometric parameters using PSO-based black-box optimization over the entire dataset. Second, the robustness modeling component embeds physical transformations into the search process: TPS~\cite{bookstein1989principal} models non-rigid spatial deformation caused by clothing wrinkles and body motion, while EOT~\cite{athalye2018synthesizing} accounts for imaging and deployment variations such as scale, translation, and sensor noise. Third, the physical deployment component materializes the optimized Curved-Block layout as wearable cold patches, enabling direct reuse without sample-specific re-optimization during deployment.

\subsection{Problem Definition}

Let the infrared pedestrian dataset be denoted as $\mathcal{J}=\{(X^{(i)},Y^{(i)})\}_{i=1}^{N}$, where $X^{(i)}\in\mathbb{R}^{H\times W}$ is an infrared image and $Y^{(i)}$ denotes the corresponding ground-truth pedestrian annotation. A pre-trained detector $f$ maps an input image to a set of detection hypotheses. For the target pedestrian hypothesis predicted by $f$ on the clean image, the detector output can be written as
\begin{equation}
  y^{(i)}=\bigl(y_{\text{pos}}^{(i)},y_{\text{obj}}^{(i)},y_{\text{cls}}^{(i)}\bigr)=f(X^{(i)}),
\end{equation}
where $y_{\text{pos}}^{(i)}$ denotes the detector-predicted bounding-box coordinates, $y_{\text{obj}}^{(i)}$ denotes objectness scores, and $y_{\text{cls}}^{(i)}$ denotes class probabilities. We use a scalar confidence functional $\phi_f(X)\in[0,1]$ to extract the objectness confidence of the target pedestrian response in image $X$. The attack therefore keeps the detector pipeline unchanged and searches for a physical patch parameter configuration that consistently suppresses $\phi_f$ after rendering, deformation, and imaging transformations.

Unlike instance-specific attacks that optimize a dedicated perturbation for each sample, UPPA learns a universal parameter configuration that is shared by all images. For a fixed grid resolution $D$ and relative perturbation width $W_p$, we represent the optimizable Curved-Block configuration by $\theta=(\delta,C)$, where $\delta=\{\delta_j\}$ collects the B\'{e}zier boundary deformation offsets and $C\in\{0,1\}^{D\times D}$ is a binary visibility matrix for the $D\times D$ Curved-Block units. Its element $c_{u,v}$ indicates whether the Curved-Block unit at grid location $(u,v)$ is activated. Here $D$ and $W_p$ control the overall perturbation scale and are selected as design hyperparameters, while $\theta$ determines the reusable shape and active topology of the universal cold-patch pattern. The admissible search domain is written as
\begin{equation}
  \begin{aligned}
  \Omega
  =
  \bigl\{
  \theta=(\delta,C)
  \mid\;& c_{u,v}\in\{0,1\},\ 1\le u,v\le D,\\
  &|\delta_j|\le \tau d,\ \forall j
  \bigr\}.
  \end{aligned}
\end{equation}
where the bound on $\delta_j$ is later derived from the topology-preserving B\'{e}zier constraint. A placement mask $\mathcal{M}^{(i)}$ is used to limit the position area of the perturbations. The adversarial sample generation process is then defined as
\begin{equation}
  X_{\text{adv}}^{(i)}(\theta)=S\bigl(X^{(i)},\theta,\mathcal{M}^{(i)}\bigr),
\end{equation}
where $S(\cdot)$ is a linear fusion function that combines the clean image with the simulated infrared Curved-Blocks inside the masked pedestrian region. The adversarial sample set induced by $\theta$ is then
\begin{equation}
  \mathcal{J}_{\text{adv}}(\theta)
  =
  \left\{
    X_{\text{adv}}^{(i)}(\theta)
    \mid
    (X^{(i)},Y^{(i)})\in\mathcal{J}
  \right\}.
\end{equation}
This formulation makes the role of the mask explicit: $\mathcal{M}^{(i)}$ restricts where the perturbation is applied, while $\theta$ carries the reusable perturbation structure shared across all images.

To reduce the digital-to-physical gap, UPPA optimizes the universal perturbation under a stochastic transformation distribution $\mathcal{T}$. Each transformation $\Gamma\sim\mathcal{T}$ combines EOT-style imaging variations~\cite{athalye2018synthesizing}, such as scaling, translation, and sensor noise, with TPS-based non-rigid deformation~\cite{bookstein1989principal} that approximates clothing wrinkles and body motion. We define the physical attack objective as the expected detector confidence after adversarial rendering and physical transformation:
\begin{equation}
  \begin{aligned}
  \mathcal{L}_{\mathrm{phys}}(\theta)
  &=
  \mathbb{E}_{X_{\mathrm{adv}}\sim\mathcal{J}_{\text{adv}}(\theta),\,\Gamma\sim\mathcal{T}}
  \left[
    \phi_f\bigl(\Gamma(X_{\mathrm{adv}})\bigr)
  \right].
  \end{aligned}
\end{equation}
The resulting constrained robust optimization problem is
\begin{equation}
  \theta^\ast
  =
  \arg\min_{\theta \in \Omega}
  \mathcal{L}_{\mathrm{phys}}(\theta).
\end{equation}
In practice, we approximate this expectation with a finite-sample average over the dataset $\mathcal{J}$ and $L$ sampled physical transformations for each image. For compact notation, we write the transformed adversarial sample as
\begin{equation}
  \tilde{X}_{\ell}^{(i)}(\theta)
  =
  \Gamma_{\ell}\bigl(X_{\mathrm{adv}}^{(i)}(\theta)\bigr),
  \quad
  \Gamma_{\ell}\sim\mathcal{T}.
\end{equation}
The empirical physical attack objective is then
\begin{equation}
  \begin{aligned}
  \widehat{\mathcal{L}}_{\mathrm{phys}}(\theta)
  &=
  \frac{1}{NL}
  \sum_{i=1}^{N}
  \sum_{\ell=1}^{L}
  \phi_f\bigl(
    \tilde{X}_{\ell}^{(i)}(\theta)
  \bigr).
  \end{aligned}
\end{equation}
This objective makes the attack explicitly universal and physically robust: the same $\theta$ is evaluated jointly over all pedestrian instances and physically plausible transformations, instead of being optimized for a single image or a single geometric state.

\subsection{Parametric Modeling of Topologically Constrained Curved-Blocks}

UPPA models the perturbation as a structured set of B\'{e}zier Curved-Blocks arranged on a $D\times D$ grid. This representation is designed for the infrared physical setting: it avoids high-frequency pixel textures, produces smooth low-temperature regions, and compresses the search space from dense pixels to a small number of geometric parameters. In practice, the fixed hyperparameters $D$ and $W_p$ determine the perturbation scale, $\delta$ controls boundary geometry, and $C$ selects the visible block topology. Together, these variables determine the curved boundary shape and the activated grid locations of the cold-patch pattern.

\subsubsection{Block Parameterization and Axial Locking}
Each activated Curved-Block unit is bounded by B\'{e}zier edges generated by the de Casteljau construction~\cite{farin2002curves}. In its general form, an $m$-th order B\'{e}zier edge is defined by a control polygon $\{P_0,P_1,\ldots,P_m\}$. Let $P_j^{(0)}=P_j$ denote the original control points. For interpolation parameter $t\in[0,1]$, de Casteljau recursion computes intermediate points as
\begin{equation}
  \begin{aligned}
  P_j^{(r)}(t)
  &=
  (1-t)P_j^{(r-1)}(t)
  +
  tP_{j+1}^{(r-1)}(t),\\
  &\qquad
  r=1,\ldots,m,\quad
  j=0,\ldots,m-r.
  \end{aligned}
\end{equation}
The final boundary point is then obtained from the last recursive level,
\begin{equation}
  B_m(t)=P_0^{(m)}(t), \quad t\in[0,1].
\end{equation}
This recursive formulation gives a unified description of B\'{e}zier boundary generation and makes the curve order explicit. In UPPA, we instantiate this general construction with $m=2$, because quadratic B\'{e}zier edges provide sufficient curvature for smooth infrared patch boundaries while keeping the number of optimizable control variables small. For one Curved-Block edge, the three control points are $\{P_0,P_1,P_2\}$. The first-level interpolants are
\begin{equation}
  \begin{aligned}
  Q_0(t) &= (1-t)P_0 + tP_1,\\
  Q_1(t) &= (1-t)P_1 + tP_2,
  \end{aligned}
\end{equation}
and the final curve point is obtained by a second interpolation,
\begin{equation}
  B(t) = (1-t)Q_0(t) + tQ_1(t).
\end{equation}
Equivalently, this recursive construction gives the quadratic B\'{e}zier form
\begin{equation}
  B(t)=(1 - t)^2 P_0+2t(1 - t)P_1+t^2P_2,\quad t\in[0,1].
\end{equation}
Here $P_0$ and $P_2$ are fixed endpoints anchored at adjacent grid vertices, while $P_1$ controls the local curvature. Instead of optimizing both coordinates of each control point, UPPA restricts the control-point displacement to a one-dimensional normal offset, which reduces the search space and preserves regular grid topology. We therefore adopt an axial locking mechanism: $P_1$ is allowed to move only along the normal direction of its corresponding grid edge,
\begin{equation}
   P_1 = P_1^{(0)} + \delta_j n, \quad n \in \{e_x,e_y\},
\end{equation}
where $P_1^{(0)}$ is the midpoint of the regular grid edge, $\delta_j\in\delta$ is the optimizable deformation offset for the $j$-th edge, and $n$ is determined by the edge orientation. For a horizontal edge, $n=e_y=[0,1]^\top$; for a vertical edge, $n=e_x=[1,0]^\top$. This one-dimensional deformation preserves the regular-grid topology while still allowing curved boundaries to adapt to human contours and clothing folds.

\subsubsection{Convex-Hull Constraint and Topological Integrity}
\begin{figure*}[t]
  \centering
  \includegraphics[width=\textwidth]{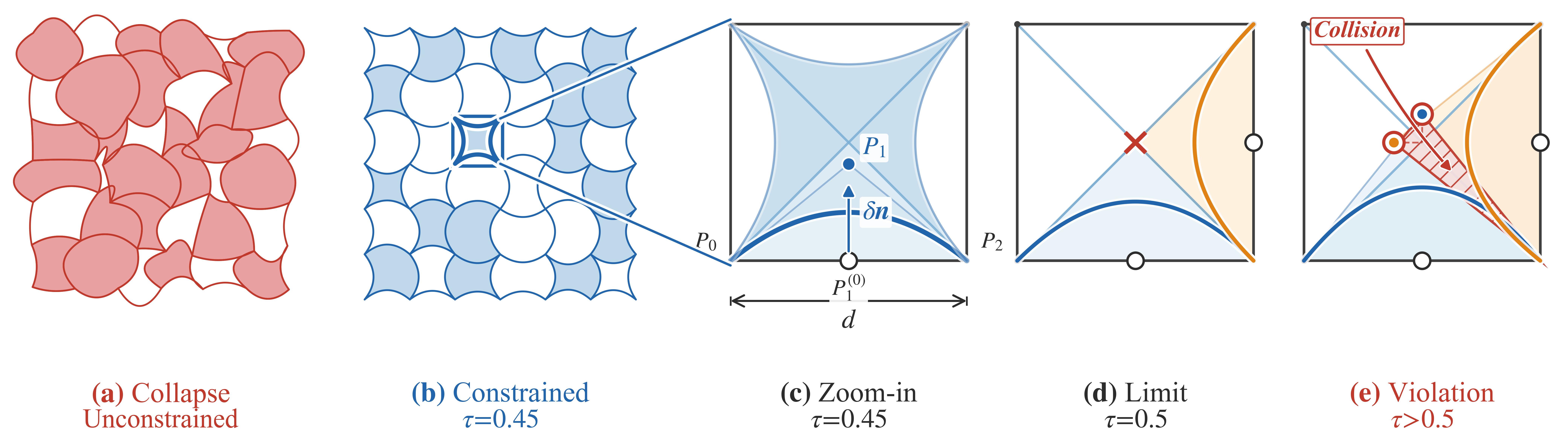}
  \caption{Topology-constrained optimization of Curved-Blocks.}
  \label{fig:topology_constraint}
\end{figure*}

Although curved boundaries improve geometric flexibility, unconstrained offsets may cause neighboring Curved-Blocks to overlap or collapse, as shown in Fig.~\ref{fig:topology_constraint}(a). In contrast, a valid Curved-Block configuration should keep adjacent curved edges separated and preserve the original grid topology, as illustrated in Fig.~\ref{fig:topology_constraint}(b). Invalid configurations not only violate physical manufacturability but also waste optimization queries in regions that cannot produce meaningful patches. To prevent this failure mode, UPPA uses the convex-hull property of B\'{e}zier curves~\cite{farin2002curves}: each curve segment remains inside the triangle formed by its control points. We partition each solid Curved-Block unit into four non-overlapping triangular territories along the diagonals, as illustrated in Fig.~\ref{fig:topology_constraint}(c), and constrain the offset of $P_1$ to remain within its assigned territory.

This construction yields a simple safety bound for the deformation parameter:
\begin{equation}
  |\delta_j| \le \tau d,
\end{equation}
where $d$ denotes the spacing of the regular grid and $\tau$ controls the allowable deformation range. The tangency limit is $0.5d$, as shown in Fig.~\ref{fig:topology_constraint}(d); once this limit is exceeded, adjacent curves may collide and break the topology, as shown in Fig.~\ref{fig:topology_constraint}(e). We therefore use the conservative bound $\tau=0.45$, which preserves a safety margin while retaining enough geometric expressiveness for smooth infrared perturbation shapes. This constraint defines the continuous deformation part of the feasible space $\Omega$. In implementation, the visibility component is binarized so that each $c_{u,v}\in C$ belongs to $\{0,1\}$ before rendering, and each updated deformation offset is mapped back to the feasible interval by the projection operator $\Pi_{\Omega}$:
\begin{equation}
  \begin{aligned}
    \Pi_{\Omega}: \quad \delta_j &\mapsto \delta_j',\\
    \delta_j'
    &=
    \operatorname{sgn}(\delta_j)
    \cdot
    \min\bigl(|\delta_j|,\tau d\bigr).
  \end{aligned}
\end{equation}
With this projection and visibility-mask binarization, every candidate perturbation remains inside the feasible space $\Omega$ and can be rendered as a stable physical Curved-Block layout.

\subsection{Black-Box Optimization via Particle Swarm Optimization}

Infrared pedestrian detectors in real deployments are typically accessed without gradients, and the Curved-Block representation contains both continuous deformation offsets and discrete visibility decisions. UPPA therefore solves the constrained universal attack objective with Particle Swarm Optimization (PSO)~\cite{kennedy1995particle}, which only requires detector confidence scores and is suitable for compact black-box search spaces. In this setting, each particle represents a complete candidate physical design: its position encodes the Curved-Block boundary offsets and activation topology under the fixed scale parameters $D$ and $W_p$, while its fitness measures how strongly this candidate design suppresses pedestrian confidence under the empirical physical-objective estimator.

For a candidate parameter configuration $\theta$, we convert the minimization of the physical attack objective into a PSO-compatible fitness maximization problem. Since $\mathcal{L}_{\mathrm{phys}}(\theta)$ is an expected detector confidence and $\phi_f(\cdot)\in[0,1]$, the fitness is defined as its complement:
\begin{equation}
  \begin{aligned}
  F(\theta)
  &=
  1-\mathcal{L}_{\mathrm{phys}}(\theta).
  \end{aligned}
\end{equation}
Thus, $F(\theta)$ is a monotone transformation of the constrained objective: maximizing $F(\theta)$ is equivalent to minimizing the detector's expected confidence on the attacked pedestrian instances. For numerical evaluation, we use the finite-sample estimator $\widehat{\mathcal{L}}_{\mathrm{phys}}$ defined above and compute the empirical fitness as
\begin{equation}
  \widehat{F}(\theta)
  =
  1-\widehat{\mathcal{L}}_{\mathrm{phys}}(\theta).
\end{equation}
This fitness design keeps all optimization feedback in the black-box confidence space: the detector is only queried for transformed adversarial samples, the returned confidence values are aggregated into $\widehat{\mathcal{L}}_{\mathrm{phys}}(\theta)$ and then converted to $\widehat{F}(\theta)$, and the swarm updates its particles according to their individual and global best fitness values.

Let $\theta_i^k$ and $v_i^k$ denote the position and velocity of the $i$-th particle at iteration $k$. The personal and global attractors are selected according to the empirical fitness:
\begin{equation}
  \begin{aligned}
  \theta_{i,\mathrm{best}}^k
  &=
  \arg\max_{\theta\in\{\theta_i^0,\ldots,\theta_i^k\}}
  \widehat{F}(\theta),
  \\
  \theta_{\mathrm{best}}^k
  &=
  \arg\max_{i}
  \widehat{F}(\theta_{i,\mathrm{best}}^k).
  \end{aligned}
\end{equation}
The PSO update then combines three terms: an inertial term preserving the current search direction, a cognitive term pulling the particle toward its personal best state $\theta_{i,\mathrm{best}}^k$, and a social term pulling it toward the global best state $\theta_{\mathrm{best}}^k$. In our implementation, $r_1$ and $r_2$ are fixed coefficients in the velocity update, with their values reported in the implementation details:
\begin{equation}
  v_i^{k+1}
  =
  \omega v_i^k
  + c_1 r_1(\theta_{i,\mathrm{best}}^k-\theta_i^k)
  + c_2 r_2(\theta_{\mathrm{best}}^k-\theta_i^k),
\end{equation}
\begin{equation}
  \theta_i^{k+1}
  =
  \Pi_{\Omega}\bigl(\theta_i^k+v_i^{k+1}\bigr).
\end{equation}

\begin{algorithm}[tb]
  \caption{UPPA optimization}
  \label{alg:uppa}
  \footnotesize
  \begin{algorithmic}[1]
    \Statex \textbf{Input:} Dataset $\mathcal{J}$, detector $f$, fixed $D$ and $W_p$, population size $N_p$, maximum iteration $K$
    \Statex \hspace{\algorithmicindent}PSO parameters $\omega,c_1,c_2,r_1,r_2$, feasible space $\Omega$, transformation distribution $\mathcal{T}$
    \Statex \textbf{Output:} Universal physical parameters $\theta^\ast$
    \State Initialize particles $\{\theta_i^0\}_{i=1}^{N_p}$ and velocities $\{v_i^0\}_{i=1}^{N_p}$
    \For{each particle $i=1,\ldots,N_p$}
      \State Project initial particle: $\theta_i^0 \leftarrow \Pi_{\Omega}(\theta_i^0)$
      \State Render Curved-Blocks to generate $\mathcal{J}_{\mathrm{adv}}(\theta_i^0)$
      \State Estimate $\widehat{\mathcal{L}}_{\mathrm{phys}}(\theta_i^0)$ under $\Gamma\sim\mathcal{T}$
      \State Compute $\widehat{F}(\theta_i^0)=1-\widehat{\mathcal{L}}_{\mathrm{phys}}(\theta_i^0)$
      \State Set personal best $\theta_{i,\mathrm{best}}\leftarrow\theta_i^0$
    \EndFor
    \State Set global best $\theta_{\mathrm{best}}\leftarrow
    \arg\max_i\widehat{F}(\theta_{i,\mathrm{best}})$
    \For{$k=0$ to $K-1$}
      \For{each particle $i=1,\ldots,N_p$}
        \State Update velocity $v_i^{k+1}$ by PSO rule
        \State $\theta_i^{k+1}\leftarrow\Pi_{\Omega}(\theta_i^k+v_i^{k+1})$
        \State Render Curved-Blocks to generate $\mathcal{J}_{\mathrm{adv}}(\theta_i^{k+1})$
        \State Estimate $\widehat{\mathcal{L}}_{\mathrm{phys}}(\theta_i^{k+1})$ under $\Gamma\sim\mathcal{T}$
        \State Compute $\widehat{F}(\theta_i^{k+1})=1-\widehat{\mathcal{L}}_{\mathrm{phys}}(\theta_i^{k+1})$
        \If{$\widehat{F}(\theta_i^{k+1})>\widehat{F}(\theta_{i,\mathrm{best}})$}
          \State $\theta_{i,\mathrm{best}}\leftarrow\theta_i^{k+1}$
        \EndIf
      \EndFor
      \State Update global best $\theta_{\mathrm{best}}\leftarrow
      \arg\max_i\widehat{F}(\theta_{i,\mathrm{best}})$
    \EndFor
    \State \Return $\theta^\ast=\theta_{\mathrm{best}}$
  \end{algorithmic}
\end{algorithm}

The projection $\Pi_{\Omega}$ enforces the topological constraints introduced above, and the visibility component is binarized before rendering, ensuring that particle updates do not leave the physically manufacturable parameter space. This projected update can be viewed as black-box search over a mixed geometric-topological feasible set: the velocity update proposes a new candidate, while $\Pi_{\Omega}$ removes deformation values that would violate the Curved-Block topology before the next fitness evaluation. We summarize the complete PSO procedure in Algorithm~\ref{alg:uppa}, including initialization, confidence-based fitness evaluation, best-state updates, and feasible-space projection after each particle update.

After each particle update, UPPA applies the projection $\Pi_{\Omega}$ to keep the candidate inside the feasible physical space, while the visibility mask $C$ determines which Curved-Block units are rendered as active cold-patch regions. This projection removes topologically invalid candidates before fitness evaluation, improving search stability and ensuring that the optimized perturbation can be manufactured as a continuous physical layout. Unlike instance-specific attacks, each candidate parameter configuration is applied to every image in the dataset and evaluated under the sampled physical transformations, encouraging the swarm to discover one shared perturbation structure that generalizes across pedestrians, scenes, and imaging conditions.

\section{Experiments}
\label{sec:experiments}

\subsection{Experimental Setting}
\label{sec:setup}

\noindent\textbf{Datasets.}
Following the experimental protocol of recent infrared physical attack studies~\cite{hu2024adversarial-blocks,tiliwalidi2025advgrid}, we use the FLIR v1\_3 dataset~\cite{flir_dataset_v1_3} for training infrared pedestrian detectors and testing digital attacks. FLIR v1\_3 contains 10,228 infrared images captured by a FLIR Tau2 thermal camera, with manually annotated objects from four categories: people, bicycles, cars, and dogs. We focus on pedestrian detection and apply a strict filtering rule that keeps only pedestrian instances taller than 120 pixels, avoiding unstable training and attack evaluation on extremely small targets. This filtering yields 1,011 pedestrian samples for detector training. The filtered FLIR v1\_3 training split is used to train the detectors, while digital attack experiments are conducted on the filtered FLIR v1\_3 test split unless otherwise specified. To test whether the learned universal perturbation generalizes beyond the source distribution, we further evaluate on four datasets that cover different physical and domain factors: FLIR v2~\cite{flir_dataset_adas} for cross-sensor imaging, LLVIP~\cite{jia2021llvip} for low-light night scenes, MFNet~\cite{ha2017mfnet} for complex backgrounds and scale variation, and M3FD~\cite{liu2022target} for diverse weather conditions.

\noindent\textbf{Object Detectors.}
We evaluate nine representative detectors from three mainstream architecture families: one-stage networks (YOLOv3~\cite{redmon2018yolov3}, RetinaNet~\cite{lin2017focal}, YOLOF~\cite{chen2021you}, YOLOX~\cite{ge2021yolox}), two-stage networks (Faster R-CNN~\cite{ren2015faster}, Mask R-CNN~\cite{he2017mask}, Libra R-CNN~\cite{pang2019libra}), and Transformer-based models (DETR~\cite{carion2020end}, Deformable-DETR~\cite{zhu2020deformable}). All detectors are trained on the curated and filtered FLIR v1\_3 training set and evaluated on the clean test set. The resulting Average Precision (AP) values for YOLOv3, DETR, Mask R-CNN, Faster R-CNN, Libra R-CNN, RetinaNet, YOLOF, YOLOX, and Deformable-DETR are 90.7\%, 91.2\%, 89.5\%, 90.8\%, 88.0\%, 93.0\%, 92.1\%, 89.3\%, and 92.8\%, respectively. These clean-set results provide a reliable basis for evaluating attack-induced degradation. Unless otherwise specified, YOLOv3 is used as the proxy model for generating universal perturbations.

\begin{figure*}[t]
  \centering
  \includegraphics[width=\textwidth]{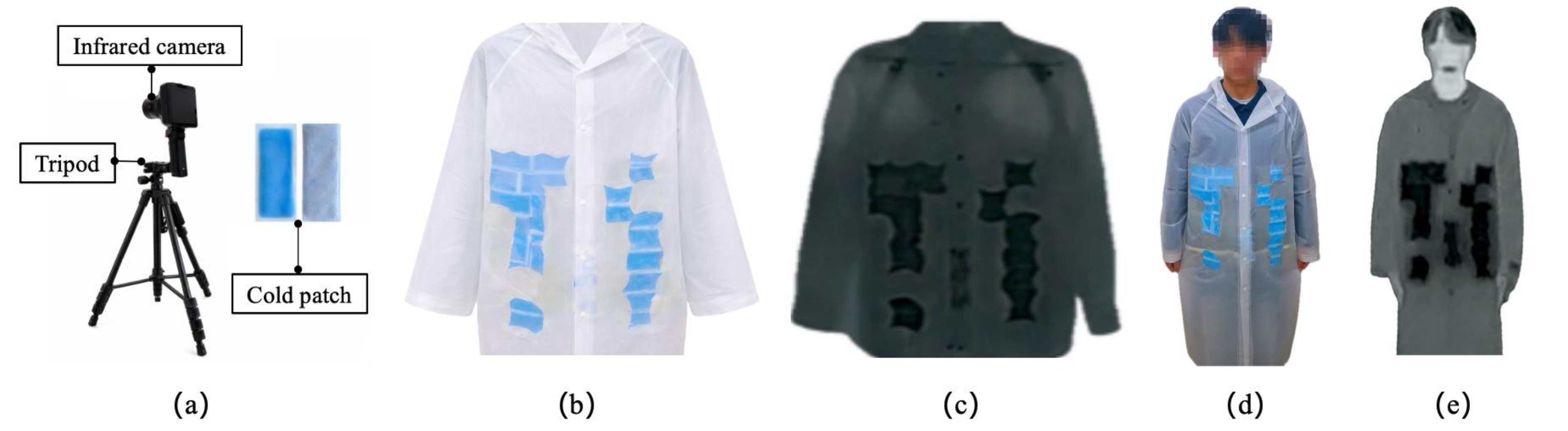}
  \caption{Experimental devices and physical deployment. (a) Infrared camera, tripod, and cold patch. (b) Visible spectrum image of the clothing. (c) Infrared image of the clothing. (d) Visible spectrum image of a pedestrian. (e) Infrared image of a pedestrian.}
  \label{fig:exp_devices}
\end{figure*}

\noindent\textbf{Experimental Devices.}
The physical setup (\mbox{Fig.~\ref{fig:exp_devices}(a)}) consists of an InfiRay XL19V2 infrared camera mounted on a tripod and wearable cold patches used as the attack carrier. The camera provides a resolution of $384 \times 288$ and a thermal sensitivity below 18\,mK, enabling stable capture of fine-grained thermal contrast in outdoor pedestrian scenes. The cold patches maintain a temperature of approximately $24\,^{\circ}\mathrm{C}$ for up to 10 hours, producing strong infrared perturbations without an external power supply. For visualization, the patches are placed inside a transparent raincoat according to the optimized B\'{e}zier layout (\mbox{Fig.~\ref{fig:exp_devices}(b)}); in practical use, they can be concealed under opaque clothing. In infrared images, the perturbations appear as smooth-bounded, low-temperature dark regions (\mbox{Fig.~\ref{fig:exp_devices}(c)}). The on-body demo (\mbox{Fig.~\ref{fig:exp_devices}(d,e)}) illustrates this physical deployment while making the patch layout visible for inspection.

\noindent\textbf{Baselines.}
We compare UPPA with HCB~\cite{wei2023hotcold}, AdvIC~\cite{hu2024adversarial-curves}, and AdvGrid~\cite{tiliwalidi2025advgrid}, which represent recent infrared black-box physical attacks based on hot--cold media, curve structures, and grid-style optimization. Because these methods are originally instance-specific, we adapt them to the same universal protocol by optimizing a shared patch over the dataset. This setting isolates the key question of whether each perturbation representation can support a reusable physical attack.

\noindent\textbf{Metrics.}
UPPA optimizes a single shared perturbation before deployment, so its detector-query budget is fixed by the PSO population size and iteration number rather than by the difficulty of individual test samples. Since no per-sample re-optimization is performed during evaluation, we use Attack Success Rate (ASR) as the primary metric. ASR is defined as the ratio of originally detected targets whose confidence falls below 0.5 after perturbation:
\begin{equation}
  \begin{aligned}
  \mathrm{ASR}
  &=
  1 -
  \frac{1}{N_t}
  \sum_{n_1=1}^{N_t}
  \mathbb{I}\bigl(y_{\text{obj}}^{n_1}\bigr),\\
  \mathbb{I}\bigl(y_{\text{obj}}^{n_1}\bigr)
  &=
  \begin{cases}
  0, & y_{\text{obj}}^{n_1} < 0.5,\\
  1, & \text{otherwise}.
  \end{cases}
  \end{aligned}
\end{equation}
where $N_t$ denotes the true positive targets detected without attacks, $y_{\text{obj}}^{n_1}$ is the predicted confidence of the $n_1$-th target under attack, and the detection threshold is 0.5.
When comparing multiple detectors on the same dataset, we additionally report the Average Attack Success Rate (AASR), which is calculated by averaging the ASR values over all evaluated infrared pedestrian detectors:
\begin{equation}
  \mathrm{AASR}
  =
  \frac{1}{N_2}
  \sum_{n_2=1}^{N_2}
  \mathrm{ASR}_{n_2},
\end{equation}
where $N_2$ is the number of infrared pedestrian detectors, and $\mathrm{ASR}_{n_2}$ represents the ASR of the $n_2$-th detector. AASR summarizes the dataset-level attack effectiveness across model architectures.

\noindent\textbf{Implementation Details.}
The default Curved-Block dimension is set to $D=6$, and the perturbation width is restricted to $W_p=1/4$ of the target bounding-box height. The topological safety threshold is set to $\tau=0.45$, following the feasible-space constraint in Section~\ref{sec3}. For PSO~\cite{kennedy1995particle}, we use population size $N_p=50$, maximum iterations $K=10$, inertia weight $\omega=0.9$, cognitive coefficient $c_1=1.6$, social coefficient $c_2=1.4$, and fixed acceleration scalars $r_1=r_2=0.5$. These settings keep the offline search budget fixed across samples, which matches the universal deployment protocol. All experiments are conducted on a single NVIDIA RTX 4090 GPU.

\begin{table*}[t]
  \centering
  \caption{Attack Success Rate (ASR) against different detection architectures and backbones across five benchmark datasets. For each detector--dataset setting, UPPA optimizes one universal perturbation over the samples in that setting. The highest ASR for each dataset is highlighted in \textbf{bold}, and the second-highest is \underline{underlined}. Models are grouped by their base architecture to illustrate relative robustness.}
  \label{tab:asr_backbone_comparison}
  \footnotesize
  \setlength{\tabcolsep}{2.5pt}
  \renewcommand{\arraystretch}{1.1}
\begin{tabular*}{0.96\textwidth}{@{\extracolsep{\fill}}lllccccc@{}}
\toprule
\multirow{2}{*}{Architecture} & \multirow{2}{*}{Target Model} & \multirow{2}{*}{Backbone} & \multicolumn{5}{c}{Datasets (ASR \%) $\uparrow$} \\
\cmidrule(lr){4-8}
 & & & FLIR v1\_3 & FLIR v2 & LLVIP & MFNet & M3FD \\
\midrule
\multirow{4}{*}{One-Stage}
 & YOLOF~\cite{chen2021you} & ResNet-50 & \underline{86.45} & 76.19 & \underline{97.14} & 93.75 & 56.83 \\
 & YOLOv3~\cite{redmon2018yolov3} & DarkNet-53& 62.42 & 47.27 & 45.45 & 86.21 & 35.66 \\
 & RetinaNet~\cite{lin2017focal} & ResNet-50 & 44.30 & 45.24 & 50.42 & 52.90 & 20.44 \\
 & YOLOX~\cite{ge2021yolox} & CSPDarknet& 42.14 & 77.16 & 53.47 & 87.02 & 77.85 \\
\midrule
\multirow{3}{*}{Two-Stage} 
 & Faster R-CNN~\cite{ren2015faster} & ResNet-50 & \textbf{89.91} & \textbf{94.78} & 91.49 & \underline{94.92} & \underline{83.74} \\
 & Mask R-CNN~\cite{he2017mask} & ResNet-50 & 85.05 & \underline{88.98} & \textbf{97.17} & \textbf{95.16} & \textbf{91.74} \\
 & Libra R-CNN~\cite{pang2019libra} & ResNet-50 & 75.00 & 80.29 & 80.19 & 84.52 & 63.64 \\
\midrule
\multirow{2}{*}{Transformer}
 & Deformable-DETR~\cite{zhu2020deformable} & ResNet-50 & 23.24 & 24.39 & 72.22 & 59.79 & 30.59 \\
 & DETR~\cite{carion2020end} & ResNet-50 & 20.20 & 36.69 & 61.54 & 46.83 & 12.84 \\
\midrule
\multicolumn{3}{l}{AASR} & 58.75 & 63.44 & 72.12 & 77.90 & 52.59 \\
\bottomrule
\end{tabular*}
\end{table*}

\begin{figure*}[t]
  \centering
  \includegraphics[width=\textwidth]{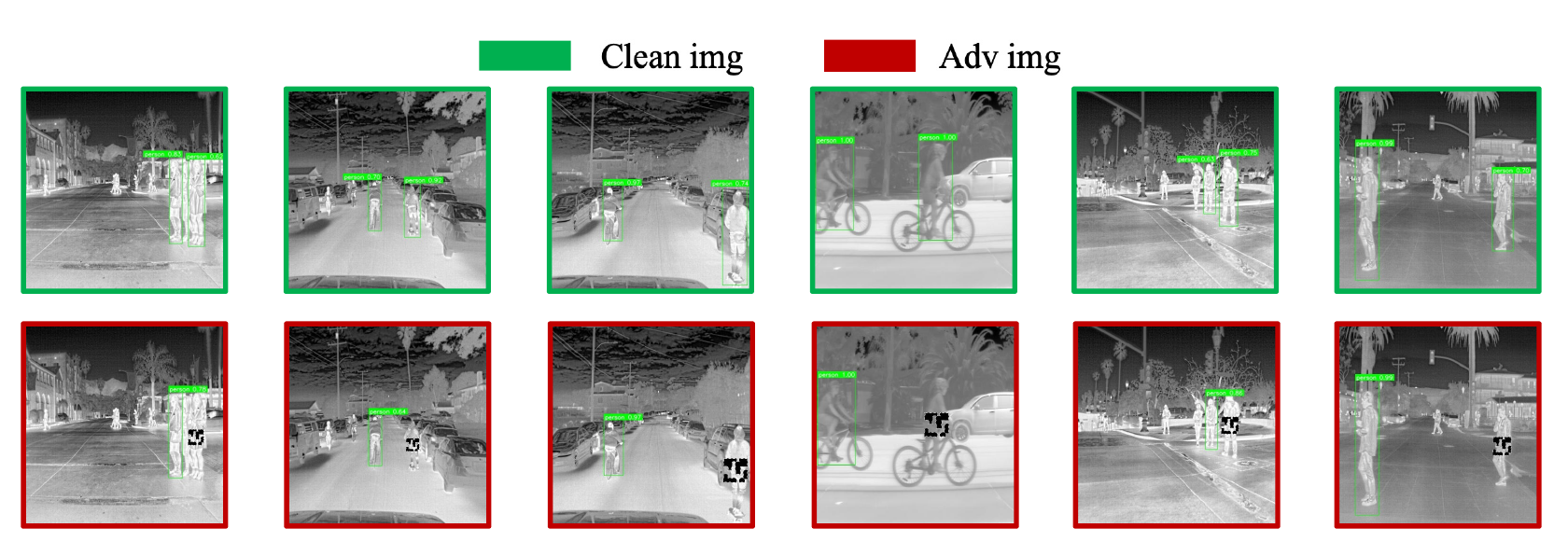}
  \caption{Digital samples generated by UPPA.}
  \label{fig:digital_samples}
\end{figure*}

\subsection{Effectiveness Evaluation}
\label{sec:effectiveness}

We first evaluate the attack effectiveness of UPPA in both digital and physical settings. The digital evaluation tests whether the Curved-Block representation can generate sample-shared perturbations for different detector--dataset settings, while the physical evaluation examines whether the digitally optimized pattern remains effective after being fabricated as wearable cold patches. This organization separates algorithmic attack effectiveness from real-world deployability.

\begin{table*}[t]
  \centering
  \caption{\protect\justifying Physical attack results of UPPA at different camera--target distances. The number of images indicates the physical samples collected at each distance, and ASR reports the attack success rate against YOLOv3.}
  \label{tab:physical_distance}
  \footnotesize
  \setlength{\tabcolsep}{8pt}
  \renewcommand{\arraystretch}{1.15}
  \begin{tabular*}{0.92\textwidth}{@{\extracolsep{\fill}}lccccccc@{}}
    \toprule
    Distance (m) & 4.8 & 5.6 & 6.0 & 6.6 & 7.2 & 7.8 & 8.4 \\
    \midrule
    Number of images & 23 & 31 & 37 & 33 & 37 & 34 & 37 \\
    ASR (\%) & 100.00 & 100.00 & 100.00 & 96.77 & 96.55 & 87.10 & 85.29 \\
    \bottomrule
  \end{tabular*}
\end{table*}

\begin{figure*}[t]
  \centering
  \includegraphics[width=\textwidth]{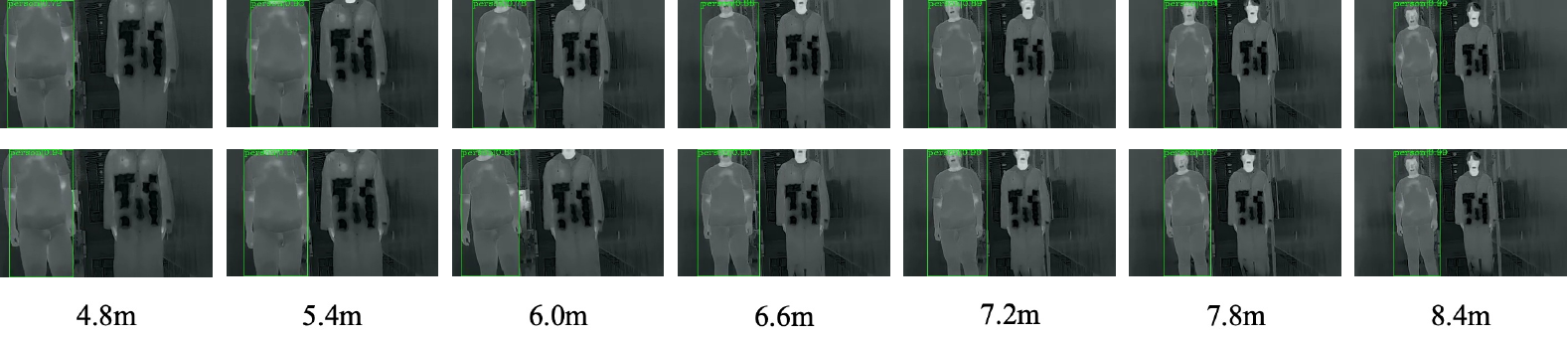}
  \caption{Representative physical samples of UPPA.}
  \label{fig:physical_exp}
\end{figure*}

\subsubsection{Digital Attacks}
We evaluate UPPA on nine detectors across five thermal datasets, with the results summarized in Table~\ref{tab:asr_backbone_comparison}. In the digital attack evaluation, a separate universal Curved-Block perturbation is optimized for each detector--dataset setting. The AASR stays above 50\% on all five benchmarks, ranging from 52.59\% on M3FD to 77.90\% on MFNet, and reaches 58.75\% on FLIR v1\_3. These results show that UPPA can find sample-shared low-temperature patterns under diverse detectors and thermal data distributions. Cross-dataset and cross-model reuse of a fixed perturbation is evaluated separately in the transferability analysis.

The attack effect is also architecture dependent. CNN-based detectors, especially two-stage models, are more vulnerable than Transformer-based detectors across most datasets. For example, Faster R-CNN and Mask R-CNN reach 89.91\% and 85.05\% ASR on FLIR v1\_3, while DETR reaches 20.20\% under the same setting. This trend is consistent with the design of UPPA: smooth B\'{e}zier Curved-Blocks disturb local thermal contours and regional feature aggregation, which are central to convolutional detection pipelines. Transformer-based detectors are less affected in several settings, suggesting that global attention can partially dilute localized thermal perturbations. However, Transformer-based detectors are not immune to UPPA; for example, Deformable-DETR still reaches 72.22\% ASR on LLVIP. Representative digital adversarial examples are provided in Fig.~\ref{fig:digital_samples}.

\subsubsection{Physical Attacks}
We further evaluate whether the digitally optimized pattern remains effective after being materialized as cold patches. Physical experiments target YOLOv3~\cite{redmon2018yolov3} at distances from 4.8\,m to 8.4\,m, covering both near-range and longer-range pedestrian observations. As shown in Table~\ref{tab:physical_distance}, UPPA reaches 100.00\% ASR from 4.8\,m to 6.0\,m and remains above 85\% even at 8.4\,m. The overall ASR is 92.59\%, showing that the optimized Curved-Block pattern survives fabrication, wearing, sensor noise, and outdoor capture. The attack gradually weakens at longer distances because the pedestrian becomes smaller in the infrared image, reducing target resolution and blurring fine thermal contrast. Representative physical samples are shown in Fig.~\ref{fig:physical_exp}, and video demonstrations are provided in the Supplementary Material.

\subsection{Stealthiness Evaluation}

\label{sec:stealthiness}

We evaluate stealthiness using both subjective ratings and an objective perceptual metric. In the subjective study, twenty volunteers rated physical sample images on a 5-point scale, yielding mean scores of 2.65, 3.20, 2.55, and 3.60 for HCB~\cite{wei2023hotcold}, AdvIC~\cite{hu2024adversarial-curves}, AdvGrid~\cite{tiliwalidi2025advgrid}, and UPPA, respectively. UPPA's smooth B\'{e}zier boundaries better mimic natural clothing folds, whereas rigid blocks and grid textures are perceived as conspicuous artifacts. We further quantify stealthiness by computing LPIPS~\cite{zhang2018unreasonable} within the local patch region localized by pixel differences. UPPA achieves the best LPIPS score (0.3667), outperforming HCB (0.3864), AdvIC (0.3724), and AdvGrid (0.4776). These results confirm that the smoothness of B\'{e}zier curves aligns the generated patterns more closely with real-world infrared thermodynamic distributions, which is important for physical deployment.

\begin{table*}[!t]
  \centering
  \caption{Cross-dataset transferability (ASR \%) of the proposed method. The perturbations are generated on the source dataset (FLIR v1\_3, shaded in gray) and evaluated on four unseen target datasets across various model architectures. Parenthesized values below each ASR denote the change relative to the source dataset. The highest transfer ASR in each target dataset is highlighted in \textbf{bold}, and the second-highest is \underline{underlined}.}
  \label{tab:cross_dataset_transfer}
  \scriptsize
  \setlength{\tabcolsep}{3pt}
  \renewcommand{\arraystretch}{1.08}
  \newcommand{\posdelta}[1]{\textcolor{green!50!black}{+#1}}
  \newcommand{\negdelta}[1]{\textcolor{red!70!black}{-#1}}
  \newcommand{\asrdelta}[2]{\begin{tabular}{@{}c@{}}#1\\[-1pt]{\tiny (#2)}\end{tabular}}
  \begin{tabular*}{0.96\textwidth}{@{\extracolsep{\fill}}llccccc@{}}
    \toprule
    \multirow{2}{*}{Architecture} & \multirow{2}{*}{Model} & \cellcolor{gray!20}Source & \multicolumn{4}{c}{Target Datasets (ASR \%) $\uparrow$} \\
    \cmidrule(lr){3-3} \cmidrule(lr){4-7}
     & & \cellcolor{gray!20}FLIR v1\_3 & FLIR v2 & LLVIP & MFNet & M3FD \\
    \midrule
    \multirow{4}{*}{One-Stage}
     & YOLOF & \cellcolor{gray!20}86.45 & \asrdelta{65.99}{\negdelta{20.46}} & \asrdelta{36.63}{\negdelta{49.82}} & \asrdelta{\textbf{95.83}}{\posdelta{9.38}} & \asrdelta{43.88}{\negdelta{42.57}} \\
     & YOLOv3 & \cellcolor{gray!20}62.42 & \asrdelta{43.64}{\negdelta{18.78}} & \asrdelta{41.41}{\negdelta{21.01}} & \asrdelta{77.01}{\posdelta{14.59}} & \asrdelta{34.27}{\negdelta{28.15}} \\
     & RetinaNet & \cellcolor{gray!20}44.30 & \asrdelta{42.26}{\negdelta{2.04}} & \asrdelta{44.54}{\posdelta{0.24}} & \asrdelta{59.42}{\posdelta{15.12}} & \asrdelta{17.52}{\negdelta{26.78}} \\
     & YOLOX & \cellcolor{gray!20}42.14 & \asrdelta{75.31}{\posdelta{33.17}} & \asrdelta{\textbf{94.29}}{\posdelta{52.15}} & \asrdelta{83.97}{\posdelta{41.83}} & \asrdelta{73.15}{\posdelta{31.01}} \\
    \midrule
    \multirow{3}{*}{Two-Stage}
     & Faster R-CNN & \cellcolor{gray!20}89.91 & \asrdelta{\textbf{87.31}}{\negdelta{2.60}} & \asrdelta{\underline{86.17}}{\negdelta{3.74}} & \asrdelta{\underline{94.92}}{\posdelta{5.01}} & \asrdelta{\underline{74.80}}{\negdelta{15.11}} \\
     & Mask R-CNN & \cellcolor{gray!20}85.05 & \asrdelta{\underline{82.68}}{\negdelta{2.37}} & \asrdelta{85.85}{\posdelta{0.80}} & \asrdelta{90.32}{\posdelta{5.27}} & \asrdelta{\textbf{76.15}}{\negdelta{8.90}} \\
     & Libra R-CNN & \cellcolor{gray!20}75.00 & \asrdelta{74.45}{\negdelta{0.55}} & \asrdelta{76.42}{\posdelta{1.42}} & \asrdelta{79.76}{\posdelta{4.76}} & \asrdelta{53.64}{\negdelta{21.36}} \\
    \midrule
    \multirow{2}{*}{Transformer}
     & Deformable-DETR & \cellcolor{gray!20}23.24 & \asrdelta{19.51}{\negdelta{3.73}} & \asrdelta{70.83}{\posdelta{47.59}} & \asrdelta{60.82}{\posdelta{37.58}} & \asrdelta{34.12}{\posdelta{10.88}} \\
     & DETR & \cellcolor{gray!20}20.20 & \asrdelta{26.62}{\posdelta{6.42}} & \asrdelta{57.26}{\posdelta{37.06}} & \asrdelta{34.92}{\posdelta{14.72}} & \asrdelta{9.17}{\negdelta{11.03}} \\
    \bottomrule
  \end{tabular*}
\end{table*}

\begin{figure}[t]
  \centering
  \includegraphics[width=\linewidth]{7.png}
  \caption{Comparison with baseline methods. (a) Dataset-wise ASR when each method is optimized on the corresponding dataset. (b) Cross-dataset transfer ASR when perturbations are generated on FLIR v1\_3 (Src.) and evaluated on FLIR v2, LLVIP, MFNet, and M3FD.}
  \label{fig:robust_models}
\end{figure}

\subsection{Comparison with Baseline Methods}
\label{sec:comparison}

As shown in Fig.~\ref{fig:robust_models}, UPPA is compared with HCB~\cite{wei2023hotcold}, AdvIC~\cite{hu2024adversarial-curves}, and AdvGrid~\cite{tiliwalidi2025advgrid} from two perspectives. The first comparison evaluates attack effectiveness when each method is optimized on the corresponding dataset (Fig.~\ref{fig:robust_models}(a)). UPPA achieves the highest ASR on all five datasets and reaches an average ASR of 55.40\%, outperforming the strongest baseline HCB by 9.98 percentage points. The advantage is especially clear on MFNet, where UPPA reaches 86.21\% ASR, suggesting that the Curved-Block representation can exploit thermal pedestrian structures more effectively than rigid blocks, curve-only patterns, or grid-style layouts. The second comparison evaluates cross-dataset transferability, where perturbations are generated on FLIR v1\_3 and directly tested on unseen target datasets (Fig.~\ref{fig:robust_models}(b)). UPPA again achieves the best average transfer ASR of 51.75\%, compared with 36.12\% for HCB, and remains effective on the challenging M3FD dataset with 34.27\% ASR. These results indicate that combining curved boundaries with a constrained block topology improves not only attack effectiveness but also the transferability of the learned thermal perturbation.
\section{Discussion}

\subsection{Transferability Evaluation}
\label{sec:transfer}

\subsubsection{Cross-Dataset Transfer}
The cross-dataset results in Table~\ref{tab:cross_dataset_transfer} examine whether UPPA learns a dataset-specific artifact or a transferable infrared vulnerability. For each detector, a single perturbation is optimized on FLIR v1\_3 and then directly deployed to FLIR v2, LLVIP, MFNet, and M3FD without fine-tuning. The perturbation transfers particularly well to MFNet, where YOLOv3 increases from 62.42\% on FLIR v1\_3 to 77.01\%, and YOLOF increases from 86.45\% to 95.83\%. This behavior suggests that UPPA is not simply memorizing FLIR v1\_3 backgrounds; instead, the learned Curved-Block pattern exploits thermal pedestrian cues that persist across datasets.

The transfer results also reveal the boundary of universality. FLIR v2 and LLVIP remain competitive for most CNN-based detectors, but M3FD is more difficult because it contains more diverse weather and acquisition conditions. This drop is visible for YOLOv3 and RetinaNet, which fall to 34.27\% and 17.52\% ASR on M3FD. Nevertheless, several detectors retain strong transfer on M3FD, such as YOLOX (73.15\%), Faster R-CNN (74.80\%), and Mask R-CNN (76.15\%), showing that the vulnerability is reduced rather than eliminated. Thus, UPPA demonstrates meaningful cross-dataset generalization, while M3FD identifies the hardest deployment regime in the current evaluation.

\begin{table*}[t]
  \centering
  \caption{\protect\justifying Cross-model transferability (ASR \%) of UPPA on FLIR v1\_3. The first column lists the source models used to generate perturbations, and the remaining columns correspond to target models. Diagonal entries report the original ASR on each source model. In each source row, the two highest transfer ASR values are marked in green and the two lowest transfer ASR values are marked in red, excluding the diagonal entry.}
  \label{tab:transferability_matrix}
  \footnotesize
  \setlength{\tabcolsep}{2.7pt}
  \renewcommand{\arraystretch}{1.16}
  \newcommand{\modelcell}[2]{\begin{tabular}[c]{@{}c@{}}#1\\[-1pt]#2\end{tabular}}
  \newcommand{\diagcell}[1]{\cellcolor{gray!18}\textbf{#1}}
  \newcommand{\highasr}[1]{\textcolor{green!55!black}{\textbf{#1}}}
  \newcommand{\lowasr}[1]{\textcolor{red!70!black}{#1}}
  \begin{tabular*}{0.98\textwidth}{@{\extracolsep{\fill}}lccccccccc@{}}
    \toprule
    \multirow{2}{*}{\textbf{Source}} & \multicolumn{9}{c}{\textbf{Target}} \\
    \cmidrule(lr){2-10}
    & \textbf{YOLOv3} & \textbf{DETR} & \modelcell{\textbf{Mask}}{\textbf{R-CNN}} & \modelcell{\textbf{Faster}}{\textbf{R-CNN}} & \modelcell{\textbf{Libra}}{\textbf{R-CNN}} & \textbf{RetinaNet} & \textbf{YOLOF} & \textbf{YOLOX} & \modelcell{\textbf{Deformable}}{\textbf{DETR}} \\
    \midrule
    YOLOv3 & \diagcell{62.42\%} & \lowasr{14.30\%} & 58.76\% & \highasr{60.20\%} & 57.73\% & 46.24\% & \highasr{63.29\%} & 35.42\% & \lowasr{26.25\%} \\
    DETR & 72.22\% & \diagcell{20.20\%} & \highasr{90.00\%} & 80.00\% & 85.00\% & 64.71\% & \highasr{91.67\%} & \lowasr{57.89\%} & \lowasr{42.86\%} \\
    Mask R-CNN & 47.06\% & \lowasr{12.09\%} & \diagcell{85.05\%} & \highasr{90.11\%} & \highasr{71.43\%} & 29.76\% & 45.21\% & 30.34\% & \lowasr{21.05\%} \\
    Faster R-CNN & 36.56\% & \lowasr{10.20\%} & \highasr{83.51\%} & \diagcell{89.91\%} & \highasr{64.95\%} & \lowasr{20.45\%} & 29.27\% & 31.18\% & 28.21\% \\
    Libra R-CNN & 56.32\% & \lowasr{12.50\%} & \highasr{81.05\%} & \highasr{86.32\%} & \diagcell{75.00\%} & 32.94\% & 56.58\% & 36.96\% & \lowasr{24.39\%} \\
    RetinaNet & 54.84\% & \lowasr{21.21\%} & 70.77\% & \highasr{77.27\%} & 70.77\% & \diagcell{44.30\%} & \highasr{80.00\%} & 34.38\% & \lowasr{26.92\%} \\
    YOLOF & 28.80\% & \lowasr{9.70\%} & 34.59\% & \highasr{43.28\%} & \highasr{39.10\%} & 26.15\% & \diagcell{86.45\%} & 22.14\% & \lowasr{10.34\%} \\
    YOLOX & 47.27\% & \lowasr{22.03\%} & \highasr{66.10\%} & \highasr{71.19\%} & 55.93\% & 39.62\% & 43.18\% & \diagcell{42.14\%} & \lowasr{32.56\%} \\
    Deformable DETR & 46.15\% & \lowasr{23.26\%} & 51.16\% & \highasr{53.49\%} & 41.86\% & \lowasr{27.50\%} & \highasr{61.76\%} & 41.86\% & \diagcell{23.24\%} \\
    \bottomrule
  \end{tabular*}
\end{table*}

\subsubsection{Cross-Model Transfer }
The cross-model results in Table~\ref{tab:transferability_matrix} examine whether perturbations optimized for one detector transfer to other model families on FLIR v1\_3. Perturbations generated on YOLOv3~\cite{redmon2018yolov3} transfer more strongly to CNN-based detectors, reaching 63.29\% ASR on YOLOF, 60.20\% on Faster R-CNN, and 58.76\% on Mask R-CNN. Transfer to Transformer-based detectors is weaker, with 26.25\% on Deformable-DETR and 14.30\% on DETR. This supports the interpretation that Curved-Blocks attack shared convolutional spatial priors rather than only a single detector head. The asymmetric transfer pattern is also informative: Transformer-to-CNN transfer can be substantial, whereas CNN-to-Transformer transfer is limited. This suggests that global attention changes how infrared pedestrian evidence is aggregated, making Transformer-based detectors less aligned with the local contour vulnerability exploited by UPPA.

\subsection{Ablation Study}
\label{sec:ablation}

\subsubsection{Ablation of the Optimization Algorithm}
We compare PSO with three alternative black-box search strategies in Table~\ref{tab:optimizer_ablation}: Random search, Genetic Algorithm (GA)~\cite{holland1992genetic}, and Differential Evolution (DE)~\cite{storn1997differential}. All variants use the same Curved-Block representation, perturbation scale, and evaluation protocol, so the comparison isolates the influence of the optimizer. Random search and GA achieve 55.41\% and 54.78\% ASR, respectively, indicating that naive sampling or genetic updates are less effective in this mixed continuous-discrete search space. DE improves the result to 61.78\%, but PSO achieves the best ASR of 62.42\%. This result supports our use of PSO: its particle-level memory and global-best guidance provide a better balance between exploration and exploitation for optimizing universal topologically constrained Curved-Blocks.

\begin{table}[t]
  \centering
  \caption{Attack performance using different optimization methods.}
  \label{tab:optimizer_ablation}
  \footnotesize
  \setlength{\tabcolsep}{7pt}
  \renewcommand{\arraystretch}{1.15}
  \begin{tabular*}{\linewidth}{@{\extracolsep{\fill}}lcccc@{}}
    \toprule
    Optimizer & Random & GA & DE & PSO (Ours) \\
    \midrule
    ASR (\%) & 55.41 & 54.78 & 61.78 & \textbf{62.42} \\
    \bottomrule
  \end{tabular*}
\end{table}

\begin{table*}[t]
  \centering
  \caption{Ablation study on different boundary formulations for perturbation generation.}
  \label{tab:ablation_boundary}
  \footnotesize
  \renewcommand{\arraystretch}{1.1}
  \setlength{\tabcolsep}{8pt}
  \begin{tabular*}{0.85\textwidth}{@{\extracolsep{\fill}}l l c@{}}
    \toprule
    \textbf{Boundary Formulation} & \textbf{Geometric Continuity} & \textbf{ASR (\%)} \\
    \midrule
    Linear & Discontinuous & 52.23 \\
    Polyline & Piecewise continuous & 53.50 \\
    Catmull-Rom spline & Smooth interpolating spline & 59.24 \\
    B\'{e}zier spline (Ours) & Flexible parametric curve & \textbf{62.42} \\
    \bottomrule
  \end{tabular*}
\end{table*}

\subsubsection{Ablation of Dimension and Width}
\begin{figure}[t]
  \centering
  \includegraphics[width=\linewidth]{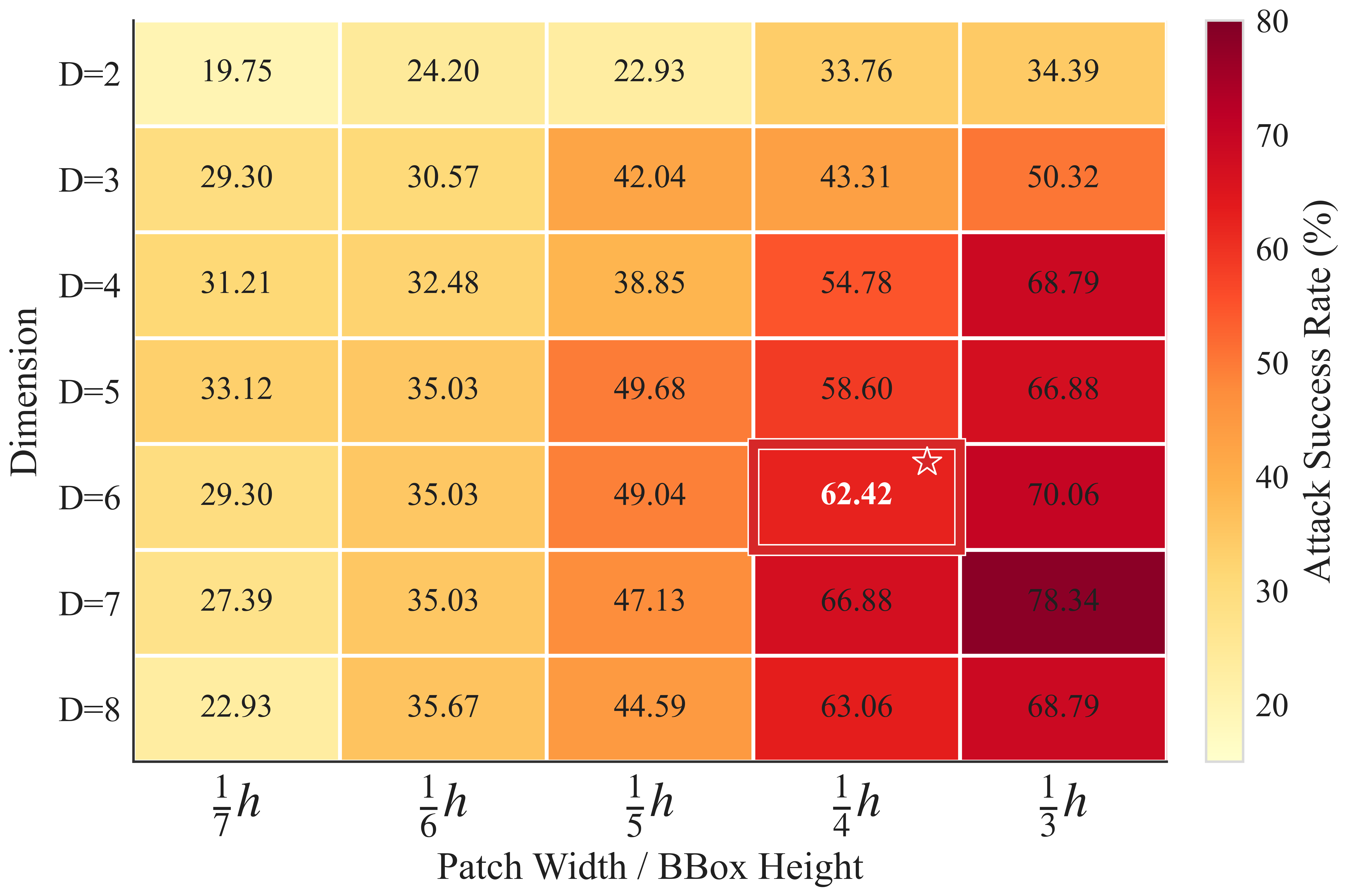}
  \caption{Ablation of Curved-Block dimension and width.}
  \label{fig:dimension_width_ablation}
\end{figure}

We study the trade-off between Curved-Block resolution $D$ and relative width $W_p$ in Fig.~\ref{fig:dimension_width_ablation}. Increasing $W_p$ generally improves ASR because wider cold regions cover more of the pedestrian's thermal signature and create stronger feature disruption. However, this gain comes with physical and perceptual costs: overly wide regions are easier to notice and less consistent with natural clothing folds. The effect of $D$ is non-monotonic: performance peaks at $D=7,W_p=1/3$ with 78.34\% ASR, but excessive fragmentation at higher resolution weakens the compact low-temperature structure needed for stable infrared attacks. We therefore use $D=6,W_p=1/4$ as the default setting because it still reaches 62.42\% ASR while offering a more balanced perturbation layout for attack strength, deployability, and visual moderation.

\subsubsection{Ablation of the B\'{e}zier Boundary Deformation Mechanism}
We isolate the contribution of the boundary representation in Table~\ref{tab:ablation_boundary}. The comparison shows that smoothness alone is not sufficient; the boundary also needs to be compact and controllable. Linear and polyline boundaries are too rigid to adapt to clothing contours, while Catmull-Rom splines improve flexibility but introduce a less constrained shape space. The proposed B\'{e}zier formulation achieves the best ASR, supporting the design choice in Section~\ref{sec3}: B\'{e}zier Curved-Blocks provide enough geometric freedom to match non-rigid pedestrian regions while retaining a compact and topologically controllable parameterization.

\subsubsection{Ablation of Grayscale Intensity}
\begin{figure*}[t]
  \centering
  \includegraphics[width=\textwidth]{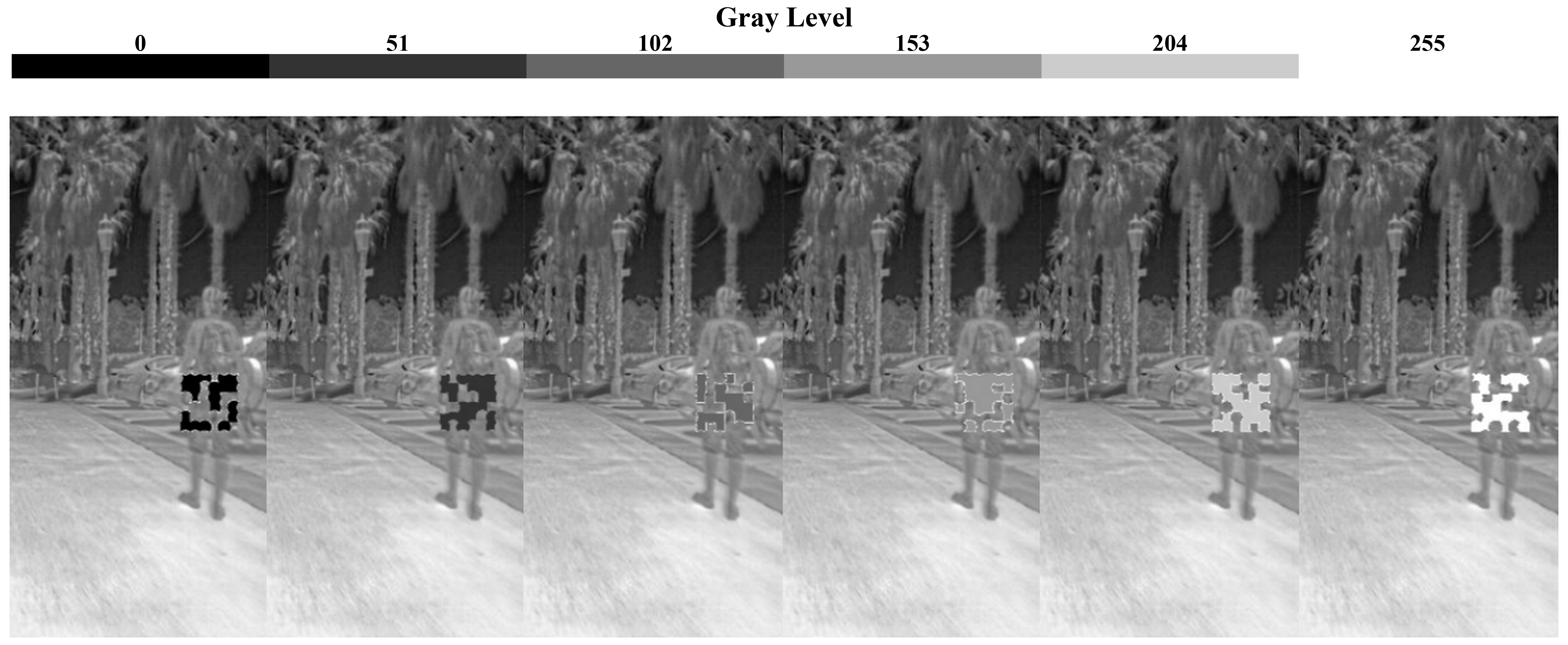}
  \caption{Ablation of grayscale level for Curved-Blocks.}
  \label{fig:color_visual}
\end{figure*}

\begin{table}[t]
  \centering
  \caption{Ablation of grayscale intensity for Curved-Blocks. Each grayscale value is applied uniformly to the three image channels in the digital proxy.}
  \label{tab:grayscale_ablation}
  \footnotesize
  \setlength{\tabcolsep}{4pt}
  \renewcommand{\arraystretch}{1.15}
  \begin{tabular*}{\linewidth}{@{\extracolsep{\fill}}lcccccc@{}}
    \toprule
    Grayscale & 0 & 51 & 102 & 153 & 204 & 255 \\
    \midrule
    ASR (\%) & \textbf{62.42} & 60.51 & 53.50 & 57.96 & 59.87 & 51.59 \\
    \bottomrule
  \end{tabular*}
\end{table}

We evaluate the effect of grayscale intensity on infrared attack strength in Fig.~\ref{fig:color_visual} and Table~\ref{tab:grayscale_ablation}. The coldest digital proxy, represented by grayscale value 0, produces the strongest attack with 62.42\% ASR, whereas the warmest proxy with grayscale value 255 gives the lowest ASR of 51.59\%. This result matches the physical motivation of UPPA: low-temperature regions create stronger contrast against warm pedestrian bodies and more effectively disrupt continuous thermal gradients. Intermediate grayscale levels fluctuate between these two endpoints but do not surpass the black setting, so grayscale value 0 is used as the digital proxy for cold physical media.

\subsubsection{Ablation of B\'{e}zier Curve Order}
\begin{figure}[t]
  \centering
  \includegraphics[width=\linewidth]{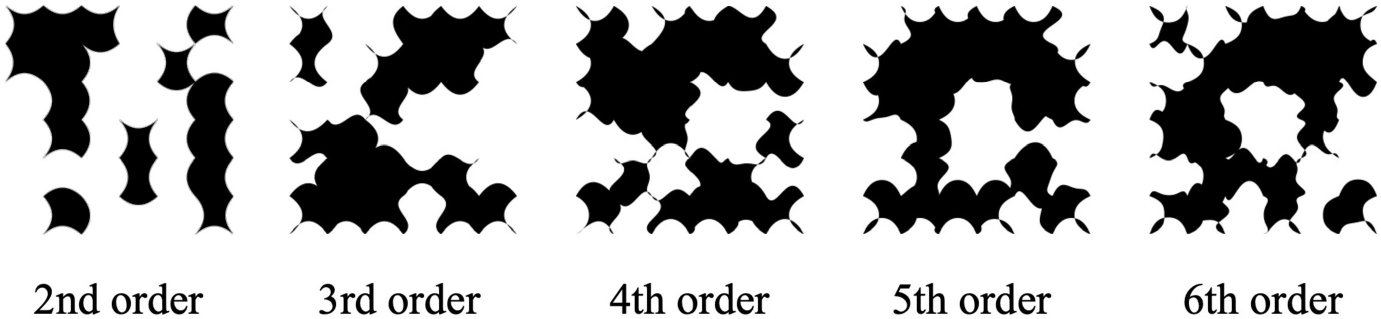}
  \caption{Visual examples of B\'{e}zier curves with different degrees from 2 to 6.}
  \label{fig:bezier_order}
\end{figure}

We further examine whether increasing the B\'{e}zier curve degree benefits the Curved-Block representation. The visual examples in Fig.~\ref{fig:bezier_order} illustrate that higher-order curves can produce more flexible boundaries, while the ASR comparison in Table~\ref{tab:bezier_order_ablation} shows that this added flexibility does not improve attack effectiveness. The 2nd-order curve achieves 62.42\% ASR, whereas variants with orders 3 to 6 drop to 40.56\%, 39.16\%, 46.85\%, and 39.86\%, respectively. This decline suggests that excessive curve freedom may introduce unstable local bends and fragmented thermal regions, weakening the coherent low-temperature contrast needed for physical infrared attacks. The 2nd-order formulation therefore provides a better balance between expressiveness, topological control, and manufacturability.

\begin{table}[t]
  \centering
  \caption{Ablation study on B\'{e}zier curve degree.}
  \label{tab:bezier_order_ablation}
  \footnotesize
  \setlength{\tabcolsep}{5pt}
  \renewcommand{\arraystretch}{1.15}
  \begin{tabular*}{\linewidth}{@{\extracolsep{\fill}}lccccc@{}}
    \toprule
    Degree & 2 & 3 & 4 & 5 & 6 \\
    \midrule
    ASR (\%) & \textbf{62.42} & 40.56 & 39.16 & 46.85 & 39.86 \\
    \bottomrule
  \end{tabular*}
\end{table}

\subsection{Visual Analysis of Attack Mechanism}
\label{sec:visual}

\begin{figure}[t]
  \centering
  \includegraphics[width=\linewidth]{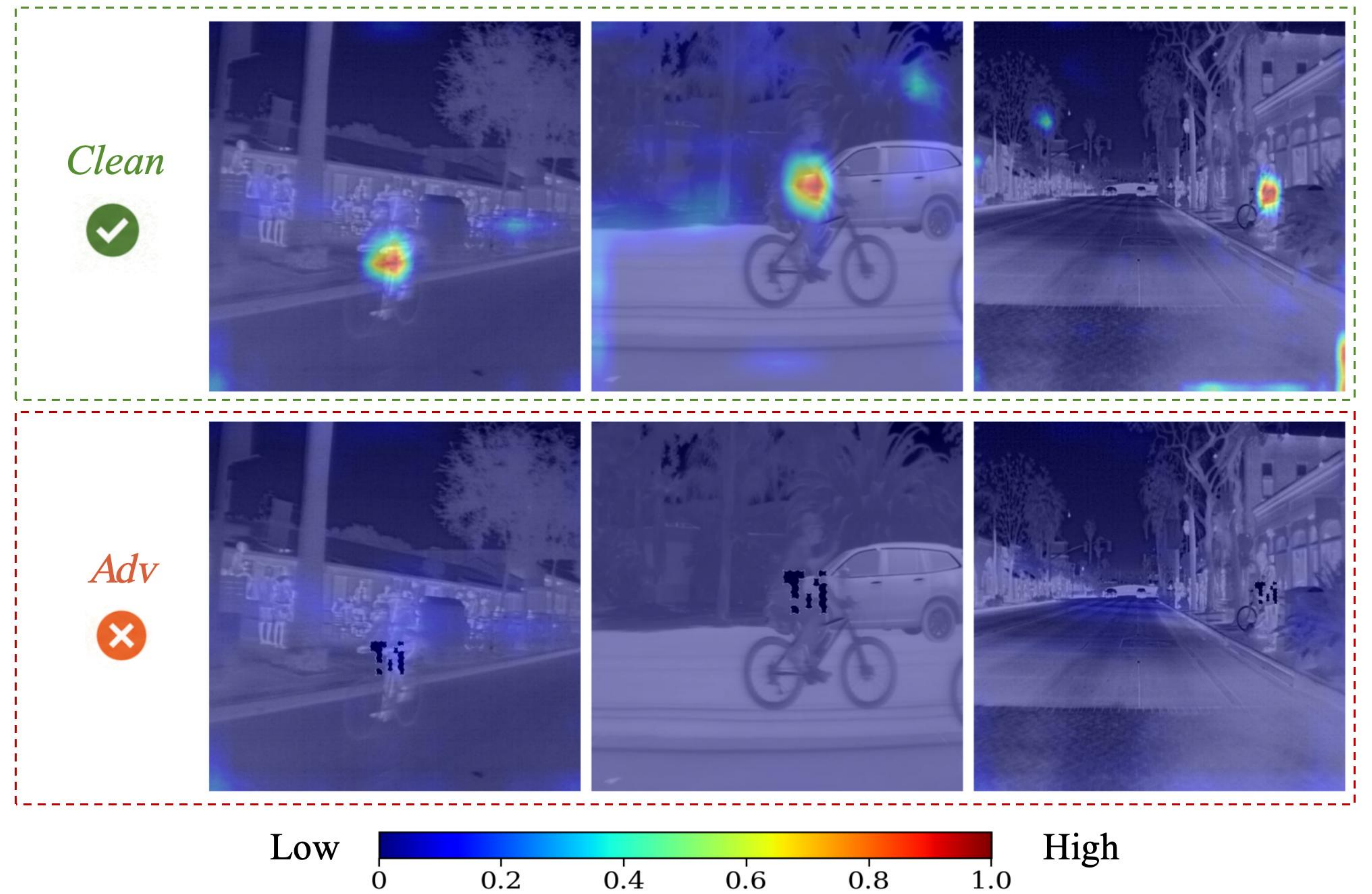}
  \caption{Grad-CAM visualization of the detector under clean and adversarial samples.}
  \label{fig:grad_cam_visual}
\end{figure}

We analyze the mechanism of UPPA in Fig.~\ref{fig:grad_cam_visual} using Grad-CAM~\cite{selvaraju2017grad} on the YOLOv3 backbone. Under the clean input, activation is concentrated on the pedestrian torso and other core regions, indicating that the detector aggregates strong target evidence from continuous thermal structures. After applying UPPA, the activation does not simply move to another object or background region; instead, it becomes globally weakened and diffuse. This observation is consistent with the quantitative results: Curved-Blocks suppress the detector's semantic evidence for the pedestrian by breaking the continuity of thermal contours and torso-level feature aggregation.

\subsection{Adversarial Defenses}
\label{sec:defenses}

To assess UPPA under practical defenses, we evaluate its performance on YOLOv3 using the MFNet dataset against two representative mechanisms: adversarial training~\cite{madry2018towards} and digital watermarking with non-blind inpainting~\cite{hayes2018visible}. Results are summarized in Table~\ref{tab:defense}.

\subsubsection{Adversarial Training (AT)}
We consider two adversarial-training settings. In AT-1, adversarial samples generated by UPPA are mixed with clean samples at a 5:1 ratio to construct the adversarial-training set, and the YOLOv3 detector is then retrained on this augmented data. The trained defense is evaluated against the optimal patch from Section~\ref{sec:effectiveness}. In AT-2, UPPA is further re-optimized against the adversarially trained detector. As shown in Table~\ref{tab:defense}, adversarial training sharply suppresses UPPA, reducing ASR from 86.21\% without defense to 28.16\% in AT-1 and 36.89\% in AT-2. The remaining 36.89\% ASR under AT-2 indicates that re-optimization can still find residual vulnerable directions, but the main conclusion is that training-based robustness substantially improves resistance to structured infrared perturbations.

\subsubsection{Digital Watermarking (DW)}
Digital watermarking with non-blind inpainting is less effective than adversarial training. UPPA retains 66.09\% ASR after this defense, suggesting that the perturbation is not a removable high-frequency artifact. Because B\'{e}zier blocks are smooth, low-frequency, and spatially coupled with the pedestrian region, inpainting can remove part of the visible perturbation but may also damage or fail to restore detection-relevant thermal cues. Therefore, restoration-style defenses are insufficient against UPPA, whereas training-based robustness is more effective in this evaluation.

\begin{table}[tb]
  \raggedright
  \caption{Evaluation of adversarial defenses against UPPA on YOLOv3.}
  \label{tab:defense}
  \footnotesize
  \renewcommand{\arraystretch}{1.2}
  \newcommand{\defdrop}[2]{\begin{tabular}{@{}c@{}}#1\\[-1pt]{\tiny \textcolor{red!70!black}{($\downarrow$ #2)}}\end{tabular}}
  \begin{tabular*}{\linewidth}{@{\extracolsep{\fill}} l c c c c @{}}
    \toprule
    & No defense & AT-1 & AT-2 & DW \\
    \midrule
    ASR (\%) & 86.21 & \defdrop{28.16}{58.05} & \defdrop{36.89}{49.32} & \defdrop{66.09}{20.12} \\
    \bottomrule
  \end{tabular*}
\end{table}

\section{Conclusion}

This paper presents UPPA, a universal physical cold-patch attack for infrared pedestrian detection. To the best of our knowledge, UPPA is the first universal physical patch attack against infrared object detectors. Unlike instance-specific or rigid-pattern attacks, UPPA uses topology-constrained B\'{e}zier Curved-Blocks to parameterize smooth low-temperature perturbations that can be optimized once and deployed without sample-specific re-optimization. This design matches the physical characteristics of infrared imaging, where thermal perturbations are naturally smooth and low-frequency rather than high-frequency visible-light textures.

Extensive evaluations across five infrared datasets and nine detectors demonstrate that UPPA is effective in both digital and physical settings. The attack shows strong cross-dataset transferability and meaningful cross-model transferability, reaches a 92.59\% ASR in real-world cold-patch experiments, and remains more effective than restoration-style defenses based on inpainting. The transfer results also reveal an architecture-dependent boundary: perturbations transfer more readily among CNN-based detectors, whereas transfer to Transformer-based detectors is harder, suggesting that global attention changes how infrared pedestrian evidence is aggregated. Ablation and visualization results further indicate that the attack works by disrupting pedestrian thermal contour continuity and regional feature aggregation. These findings reveal a practical universal physical vulnerability in current infrared pedestrian detectors while clarifying where this vulnerability is less easily transferred.

The current study is limited to fixed universal patterns for infrared pedestrian detection, and performance is weaker under harder transfer regimes, including CNN-to-Transformer transfer and challenging domain shifts such as M3FD. Future work should extend universal infrared attacks to dynamic, multi-person, and multimodal perception scenarios, while developing defenses that combine adversarial training with physics-aware infrared restoration.

\backmatter

\bmhead{Data availability}

The datasets used and/or analysed during the current study are publicly available from the sources cited in the manuscript. Additional data generated during this study are available from the corresponding author on reasonable request.

\bibliography{sn-bibliography}

\end{document}